\documentclass[journal]{IEEEtran}
\usepackage{times}
\usepackage{epsfig}
\usepackage{graphicx}
\usepackage{amsmath}
\usepackage{amssymb}
\usepackage{adjustbox}
\usepackage{subfig}
\usepackage{bm}
\usepackage{acronym}
\usepackage{adjustbox}
\usepackage[table]{xcolor}
\usepackage{svg}
\usepackage{mathtools}
\usepackage{bbm}
\usepackage{float}
\usepackage{tikz}
\usepackage{dblfloatfix}
\definecolor{jungle}{HTML}{00A99A}

\newcommand\datasetname{voraus-AD}

\newcommand\anoImgWidth{0.16}
\newcommand\imgSpaceBetween{0.1mm}
\newenvironment{packed_enum}{
\begin{itemize}
  \setlength{\itemsep}{3pt}
  \setlength{\parskip}{0pt}
  \setlength{\parsep}{0pt}
}{\end{itemize}}

\usepackage{pifont}
\newcommand{\xmark}{\ding{55}}
\usepackage{multirow,makecell, rotating}
\usepackage{booktabs}
\usepackage{tikz}
\usepackage{graphbox}
\usepackage{comment}
\usepackage{epsfig}
\usepackage{bm}
\usepackage{soul} 
\usepackage{color}
\usepackage{verbatim}
\usepackage{etoolbox}
\usepackage{mathtools}
\usepackage{tabularx}
\usepackage{wasysym}
\usepackage{lipsum}

\newcommand\blfootnote[1]{%
  \begingroup
  \renewcommand\thefootnote{}\footnote{#1}%
  \addtocounter{footnote}{-1}%
  \endgroup
}

\usepackage{multirow,makecell, rotating}

\newcolumntype{Y}{>{\centering\arraybackslash}X}
\DeclarePairedDelimiter\abs{\lvert}{\rvert}%
\makeatletter
\let\oldabs\abs
\def\abs{\@ifstar{\oldabs}{\oldabs*}}
\DeclarePairedDelimiter\norm{\lVert}{\rVert}%
\let\oldnorm\norm
\def\norm{\@ifstar{\oldnorm}{\oldnorm*}}
\makeatother
\usepackage[pagebackref=true,breaklinks=true,letterpaper=true,colorlinks,bookmarks=false]{hyperref}

\definecolor{lime}{HTML}{A6CE39}
\DeclareRobustCommand{\orcidicon}{
	\hspace{-3mm}
	\begin{tikzpicture}
		\draw[lime, fill=lime] (0,0) 
		circle [radius=0.16] 
		node[white] {{\fontfamily{qag}\selectfont \tiny ID}};
		\draw[white, fill=white] (-0.0625,0.095) 
		circle [radius=0.007];
	\end{tikzpicture}
	\hspace{-2mm}
}
\foreach \x in {A, ..., Z}{%
	\expandafter\xdef\csname orcid\x\endcsname{\noexpand\href{https://orcid.org/\csname orcidauthor\x\endcsname}{\noexpand\orcidicon}}
}

\author{
	Jan Thieß Brockmann\textsuperscript{*}\orcidA{}, 
	Marco Rudolph\textsuperscript{*}\orcidB{}, 
	Bodo Rosenhahn\orcidC{}, 
	Bastian Wandt\orcidD{}
\thanks{J. T. Brockmann is with voraus robotik GmbH, 30167 Hanover, Germany~(mail: thiess.brockmann@vorausrobotik.com).

	M. Rudolph and B. Rosenhahn are with the Institute of Information
Processing, L3S, Leibniz University Hannover, 30167 Hannover, Germany (mail: \{rudolph, rosenhahn\}@tnt.uni-hannover.de).

	B. Wandt is with the Computer Vision Laboratory, Linköping University, SE-581 83 Linköping, Sweden
~(mail: bastian.wandt@liu.se).}
	\thanks{
	This work was supported by the Federal Ministry of Education and Research (BMBF), Germany under the the AI service center KISSKI (grant no. 01IS22093C), the Deutsche Forschungsgemeinschaft (DFG) under Germany’s Excellence Strategy within the Cluster of Excellence PhoenixD (EXC 2122) and the German Federal Ministry of the Environment, Nature Conservation, Nuclear Safety and Consumer Protection (GreenAutoML4FAS project no. 67KI32007A).
}
}

\begin{document}

\title{The voraus-AD Dataset for Anomaly Detection\\ in Robot Applications}

\markboth{}
{Shell \MakeLowercase{\textit{et al.}}: Bare Demo of IEEEtran.cls for Journals}

\maketitle

\begin{abstract}
During the operation of industrial robots, unusual events may endanger the safety of humans and the quality of production.
When collecting data to detect such cases, it is not ensured that data from all potentially occurring errors is included as unforeseeable events may happen over time.
Therefore, anomaly detection (AD) delivers a practical solution, using only normal data to learn to detect unusual events.
We introduce a dataset that allows training and benchmarking of anomaly detection methods for robotic applications based on machine data which will be made publicly available to the research community.
As a typical robot task the dataset includes a pick-and-place application which involves movement, actions of the end effector and interactions with the objects of the environment.
Since several of the contained anomalies are not task-specific but general, evaluations on our dataset {are transferable to} other robotics applications as well.
Additionally, we present \textit{MVT-Flow} (multivariate time-series flow) as a new baseline method for anomaly detection:
It relies on deep-learning-based density estimation with normalizing flows, tailored to the data domain by taking its structure into account for the architecture.
Our evaluation shows that MVT-Flow outperforms baselines from previous work by a large margin {of 6.2\% in area under ROC}.

\end{abstract}

\maketitle
\section{Introduction}
\blfootnote{* denotes an equal contribution}
\IEEEPARstart{A}{s} a subfield of automation, the use of robotic arms is essential for industrial applications and processes~\cite{bahrin2016industry}.
Industrial robots can relieve humans of monotonous and dangerous tasks and perform them with consistent quality without interruption.
With automated monitoring, robotic arms can operate continuously without the presence of humans.
Detecting unusual behavior during automated execution of robotic applications is critical for ensuring process and product quality as well as for human safety~\cite{vasic2013safety}.
With the increasing importance of collaboration (e.g., human-machine interactions) and the growing use of artificial intelligence, the complexity of robotic applications is increasing as well compared to the more monotonous applications with traditional industrial robots~\cite{kragic2018interactive}.
For example, objects to be gripped no longer have to be in a fixed position, but are localized by visual object recognition.
Since naive approaches are no longer feasible for the detection of unusual events, the design of more sophisticated machine learning methods is coming more and more into the interest of research~\cite{hornung2014model, swvae, rnn_vae, romeres2019anomaly, khalastchi2015online, zhang2021robot}.
Another challenge for the detection is the availability of data:
At the beginning of a new deployment, there are no unusual events that could be used to train a machine-learning-based classifier.
Even though a small amount of data from anomalous events, which are usually rare, may be available, it can never be ensured that all scenarios are represented in the collected data as machine wear, unpredictable changes in the environment or chain reactions over several process steps may happen in the future~\cite{mvtec}.
This makes a traditional supervised classification impractical.
Instead, the detection of unusual behavior requires semi-supervised anomaly detection (AD): 
Using only \textit{normal data} without any anomaly for training, it should be detected whether a given example is normal or anomalous, i.e. an unusual event in the application.
Note that this setting is sometimes referred to in the literature as unsupervised anomaly detection~\cite{mvtec3d, st_bergmann2, memae}\footnote{In the original definition, the training set in unsupervised AD may contain anomalies without having labels for that. Most of the literature focus on the semi-supervised case.}.
\begin{figure}
  \centering
  \includegraphics[width=0.96\linewidth]{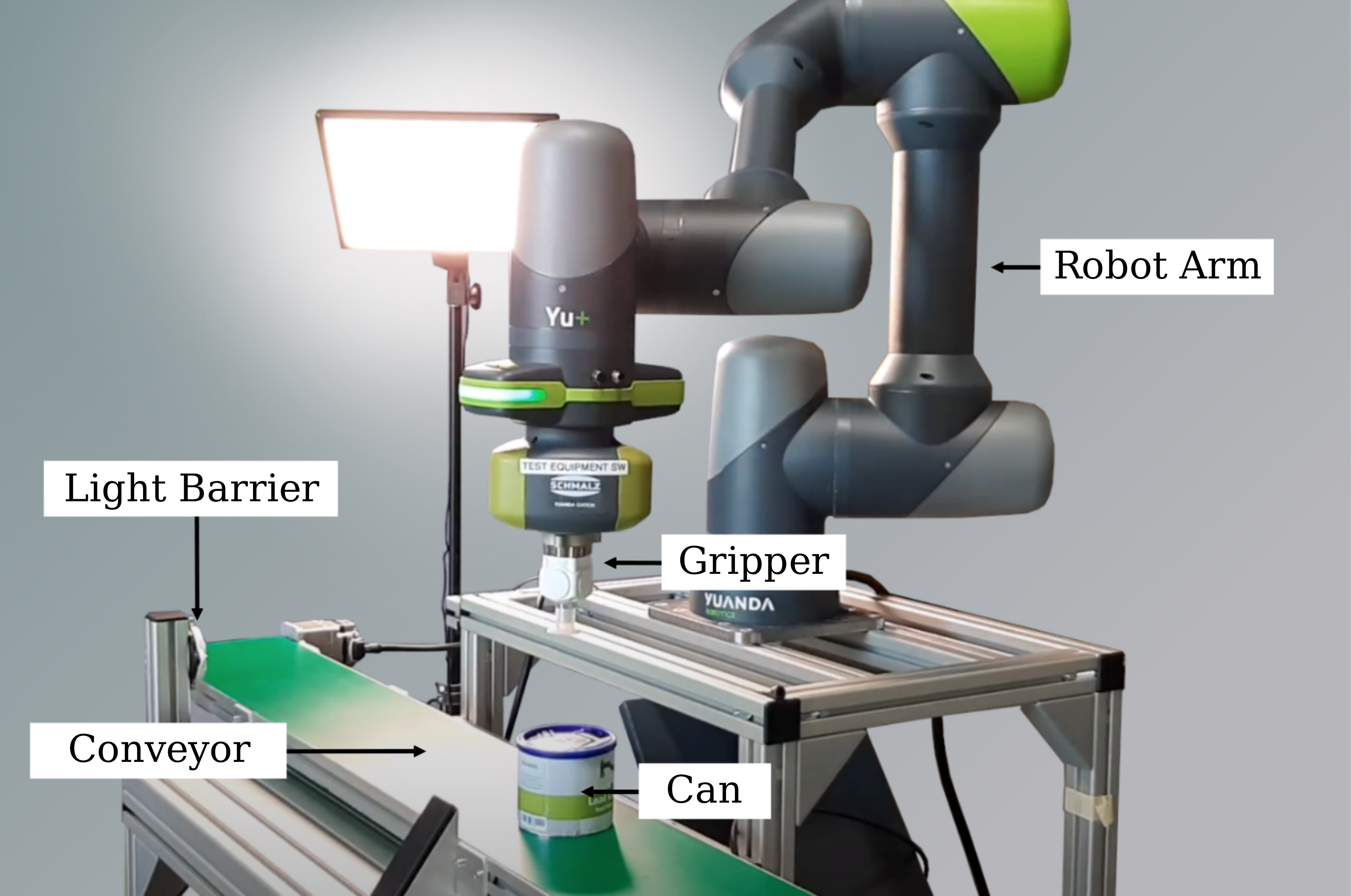}
  \caption{Setup of the pick-and-place application in voraus-AD: A {robotic arm} grips a randomly positioned can and moves it to a fixed position.}
  \label{fig:setup}
  \vspace{-5mm}
\end{figure}

In order to ensure comparability of different methods w.r.t. different applications and to establish transparency, it is common to evaluate on public datasets.
Unfortunately, this is hardly the case for AD in robot applications at the moment.
The only public dataset \textit{AURSAD}~\cite{aursad} is not widely used since {it is} very specific to a screwdriver application.
Publications {have so far been} evaluated on private datasets~\cite{hornung2014model, swvae, rnn_vae, romeres2019anomaly, khalastchi2015online, zhang2021robot} which makes the experiments irreproducible.
This issue motivates us to present a publicly available benchmark based on machine data which enables an appropriate evaluation due to the included number, types, and diversity of anomalies.
As an application, we choose a standard pick-and-place operation, whose setup is shown in Figure~\ref{fig:setup}, that includes typical elements such as the movement of a {robotic arm}, actions of the end effector, and interaction with the environment.
The dataset contains 1367 {time series, which we will call \textit{samples} in the following}, of anomaly-free data, in which a  {collaborative} robot picks a randomly placed can with a vacuum gripper and places it at a pre-determined position.
For the test set, 755 additional samples, {divided in} 12 anomaly types, were intentionally induced, which include robot axis wear, gripping errors and process errors such as collisions, manipulation of the can or an unstable setup.
In contrast to AURSAD, where the anomalies are specific to its screwdriver application, many of the anomalies in our dataset, such as axis wear or collisions, also occur in other applications. 
Thus, the evaluation on our dataset provides a more general validity about the respective method.
We define an evaluation protocol that individually evaluates each anomaly type.

Since the usually integrated safety systems of {collaborative} robots interrupt the operation in case of large deviations from the norm, which can be detected easily, we especially chose subtle anomalies, which would not be recognized by the safety systems.
For example, our dataset includes collisions with a free-hanging cable, which impacts the signals only marginally, although tangling of this cable may be a major safety hazard.
In addition, different intensities of anomalies were simulated in each case, for example by varying the weight of the can in several steps.

Since the machine data of the robot is usually recorded, we model a practicable use case in that no additional sensing is required.
The signals include a total of 78 mechanical signals as well as 52 electrical signals from a total of 6 axes at a sampling frequency of 500 Hz.
{Furthermore}, we provide metadata which, for example, gives detailed information about the anomaly and the status of the robot.

{We observed that current techniques struggle} with the processing of the large number and the different characteristics of the provided signals.
Therefore, in addition to the dataset, we present a new AD baseline method \textit{MVT-Flow} for multivariate time series which is based on estimating the probability density of the normal data {and inspired from works in other domains~\cite{csflow, kang2022traffic}}.
It is assumed that samples with low likelihood represent an anomaly.
To model the density, we use a deep-learning-based normalizing flow (NF)~\cite{nf} to learn a transformation between the data distribution and a normal distribution in which likelihood {is} measurable.
While NFs are already established in anomaly detection on images~\cite{differnet, csflow, cflow, yan2022cainnflow}, we adapt them to the different data structure of multivariate time series.
So far, NFs are used to transform one-dimensional time series with fully connected layers as internal networks as in~\cite{nf_time_series}, which is not applicable to the high dimensionality of extensive machine data.
{Dai and Chen circumvent this problem by using a Bayesian network to combine likelihoods of single NFs each processing one signal~\cite{GANF}.}
{We propose an NF architecture that allows multivariate time series to be processed with a single network} by exploiting the temporal structure via convolutions in the internal subnetworks of the model while preserving the concept of signals.
Unlike other density-based AD methods~\cite{romeres2019anomaly, khalastchi2015online}, our learned density estimation is significantly more flexible to data distributions with no strict assumptions to be made.
Additionally, MVT-Flow enables a temporal analysis to identify when an unusual event occurs.

Summarizing, our \textbf{contributions} are as follows:
\begin{packed_enum}
    \item We publish a comprehensive anomaly detection dataset based on machine data from a robot, which includes regular passes as well as anomalies during a pick-and-place task in a realistic setup.
    \item Our challenging dataset contains a large diversity of anomalies in terms of error source and intensity which are both task-specific and more general.
    \item We propose an AD method by tailoring Normalizing Flows to multivariate time series and outperform existing methods on our dataset by a large margin. The code and dataset is available at github\footnote{\url{https://www.tnt.uni-hannover.de/vorausAD}}.
\end{packed_enum}

\section{Related Work}
In this section, we describe methods as well as datasets of previous work on anomaly detection for robot applications in Sections~\ref{ad_methods} and \ref{ad_datasets}.
For a general overview of anomaly detection methods and their usage in other data domains, we refer the reader to~\cite{ad_review1, ad_survey1, ad_survey2}.
In Section~\ref{nf_related} we introduce Normalizing Flows, on which our baseline is based, and related work that applies NFs to anomaly detection.

\subsection{AD Methods for Robot Applications}
\label{ad_methods}
In the following, we provide an overview of past work regarding AD methods in the context of robotic applications.
Note that we focus on semi-supervised anomaly detection on machine data where no anomalies are given in the training set and no external sensors are used.
{A broader overview of anomaly detection methods in robotics is given in~\cite{khalastchi2018fault}.}

Hornung et al.~\cite{hornung2014model} identify anomalies by performing a Principal Component Analysis (PCA~\cite{pca}) on normal data and use the reconstruction error as an anomaly indicator.
It is assumed that anomalies are poorly reconstructed due to increased variance on axes that are not modeled by the PCA.
Chen et al.~\cite{swvae} and Sölch et al.~\cite{rnn_vae} train adapted versions of variational autoencoders~\cite{vae} to detect anomalies using the reconstruction error similarly to the PCA-based concept.
This should be larger for anomalies than for normal data since it is only trained on the latter.
While~\cite{swvae} is based on convolutional layers,~\cite{rnn_vae} integrates residual neural networks (RNNs).
Park et al.~\cite{park2018multimodal} {also makes} use of RNN-based VAEs by integrating LSTMs, however, it utilizes the evidence lower bound (ELBO) as the anomaly score.
Learning a progress-based latent space that can be partitioned by only a small set of nominal observations but is also built with unlabeled examples is proposed by Azzalini et al.~\cite{azzalini2021minimally}.

Addressing the problem from a statistical perspective,  Khalastchi et al. \cite{khalastchi2015online} and Romeres et al.~\cite{romeres2019anomaly} try to model the distribution of normal samples.
Khalastchi et al.~\cite{khalastchi2015online} apply the Mahalanobis distance within a sliding window as an anomaly score which is equivalent to computing the negative log likelihood of a multivariate Gaussian.
However, the assumption of the data distribution being Gaussian is a strong simplification and very inflexible to more complex distributions.
Romeres et al.~\cite{romeres2019anomaly} evaluate Gaussian process models and some extensions of these.
Gaussian process models require to find appropriate kernels so that the assumption of a Gaussian process is valid.
In contrast, our baseline also models the distribution of normal samples but is capable of handling any data distribution without any manual kernel search.

Zhang et al.~\cite{zhang2021robot} analyze the time series with an autoregressive model that predicts future torques.
The error between this prediction and the actual signals should identify anomalies for real-time collision detection.
Again, it relies on linear models that cannot model more complex mechanisms.

Park et al.~\cite{park2016multimodal} use Hidden Markov Models (HMMs~\cite{hmm}) by utilizing the observation probability as an indicator for unusual events after fitting a HMM to normal data, which was further enhanced with Gaussian process regression~\cite{park2019multimodal}.
Azzalini et al.~\cite{hmm_ad} propose two alternative approaches for HMMs: The online approach uses the Hellinger distance~\cite{hellinger1909neue} to calculate the emission distribution between normal data and test data in the present state.
In an offline version, a separate HMM is fitted on test data and the states are matched to those of the normal data, {before applying the Hellinger distance between these models as an anomaly score}.

\subsection{AD Datasets for Robot Applications}
\label{ad_datasets}
Unfortunately, there is a lack of public datasets in the literature for anomaly detection in robotic applications.
Most work evaluates on private datasets.
The only publicly available dataset \textit{AURSAD}~\cite{aursad} contains the machine data of a robot during the execution of a screwdriver application.
The robotic arm used is of the UR3e type from \textit{Universal Robots} and has 6 axes arranged in series.
For the application, the robotic arm is mounted on a table with two perforated plates with 92 threaded holes each.
The total of 125 signals include the angles, velocities and currents of the axis motors. In addition, the total voltage and current of the robotic arm, other target values, and the machine data of the screwdriver are part of the dataset.

The robotic arm is equipped with an electric screwdriver on the end effector. 
During the execution of the application, the robotic arm first moves from the starting position over a screw on one plate which is then unscrewed.
The robot moves to a threaded hole in the other plate.
This process is performed repeatedly.
In addition to the normal data, the dataset contains recordings of only four different anomalies.
These include recordings of screwdriving processes with defective screws, additional assembly components and missing screws.
For the fourth anomaly, a previously demolished plate thread, only three samples are given.
Each of these four anomalies is very specific to the screwdriver application and, in contrast to many anomalies in our dataset, may not occur in other applications.
A detailed comparison and differentiation between AURSAD and our dataset are given in Sec.~\ref{dataset_comparison}.
\begin{figure}
\centering
\resizebox{1\linewidth}{!}{
\begin{tabular}{ccc}
Data Space $\mathcal{X}$ & & Latent Space $\mathcal{Z}$ \\
\raisebox{-.43\height}{
\includegraphics[width=0.4\linewidth]{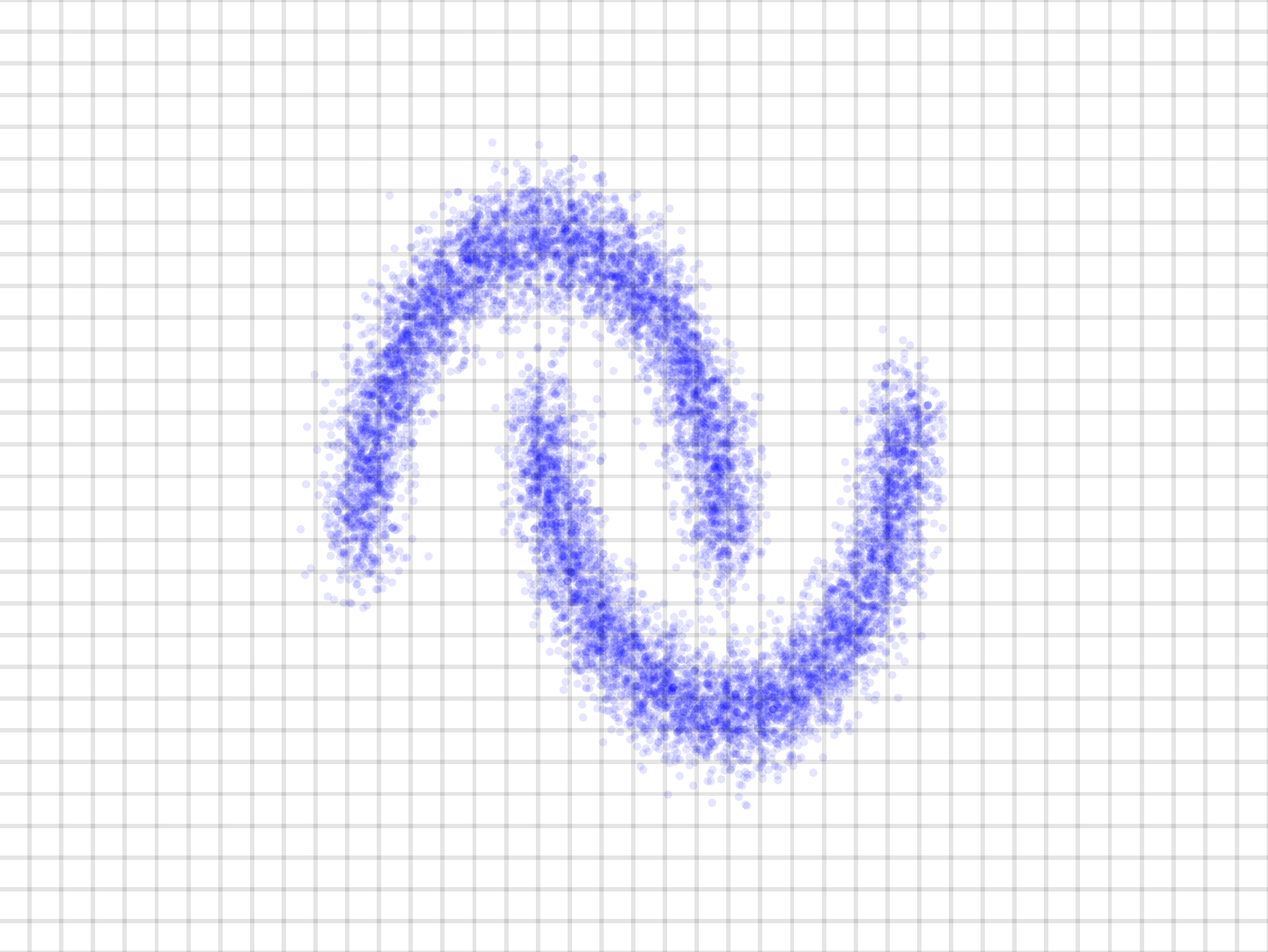}
}
&
\hspace{-5mm}
\begin{tabular}{c}
\textbf{NF}\\
$\Leftrightarrow$\\
$z = f(x)$ \\
\end{tabular}
\hspace{-5mm}
&
\raisebox{-.43\height}{
\includegraphics[width=0.4\linewidth]{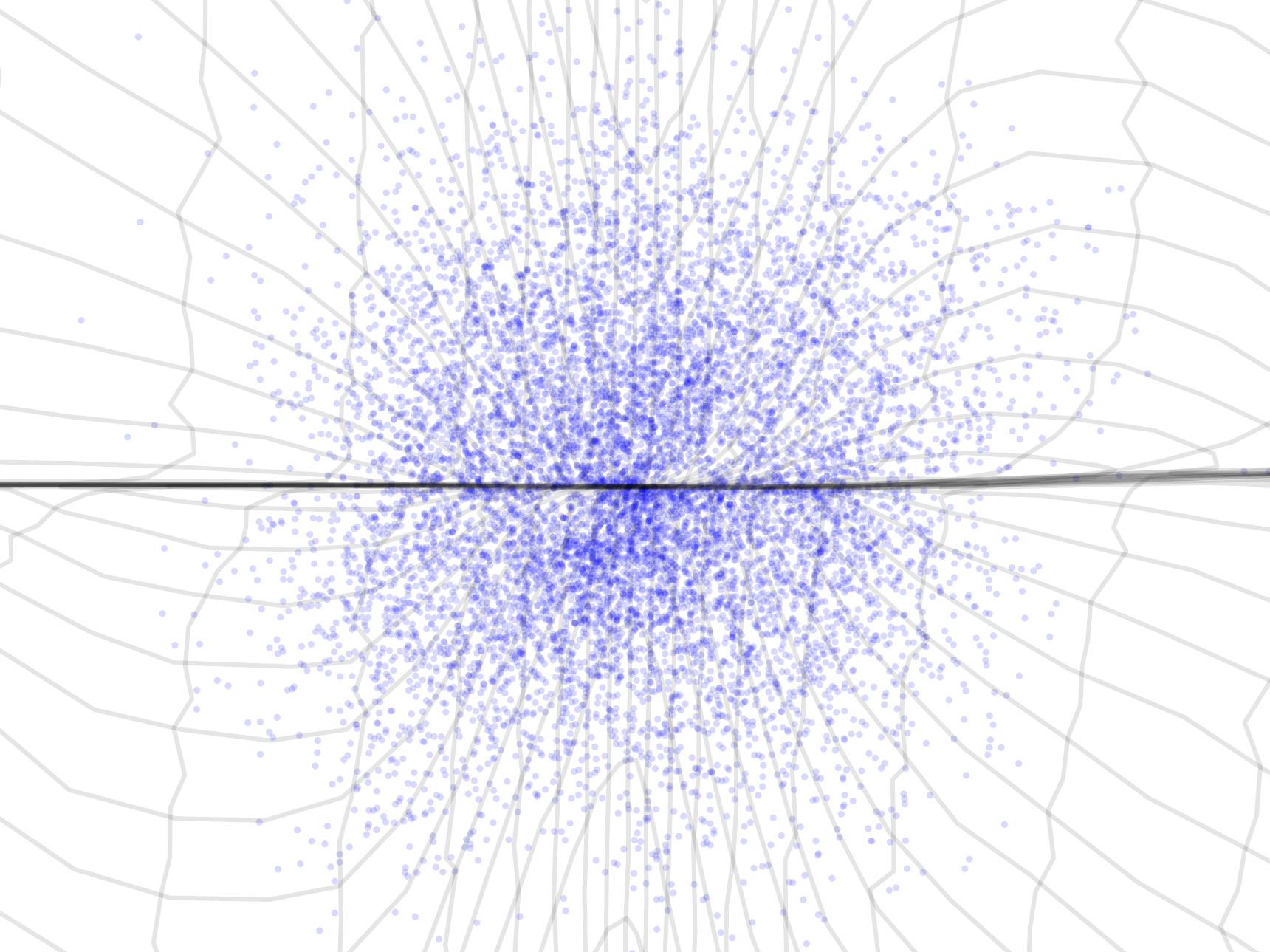}
}
\\
$x \sim p_{X}$ & & $z\sim p_{Z}$ with $p_{Z} = \mathcal{N}(0,\,I)$
\end{tabular}}
    \caption{A Normalizing Flow learns a transformation between an unknown data distribution to a well-defined distribution, which is typically a Gaussian. Images are taken from~\cite{realnvp}.
    }
    \label{fig:nf_transform}
\end{figure}

\subsection{Normalizing Flows}
\label{nf_related}
A Normalizing Flow, introduced by Rezende and Mohamed~\cite{nf}, is a generative neural network that is mostly used to map between a data distribution~$X$ and a well-defined distribution~$Z$ as visualized in Figure~\ref{fig:nf_transform}.
Normalizing Flows differ from other generative models such as VAEs~\cite{vae} or GANs~\cite{gan} in that they map bijectively and are bidirectionally executable.
Usually, they are optimized via maximum likelihood training, which is described in detail in Section~\ref{training}, to estimate the density of the data distribution.
Subsequently to the training process, they can be used in two ways:
The forward pass $f:\mathcal{X}\rightarrow \mathcal{Z}$ allows for computing the likelihood of data points given the well-defined distribution $p_Z$.
Conversely, the backward pass $f^{-1}:\mathcal{Z}\rightarrow \mathcal{X}$ allows for generating new samples in the original space $\mathcal{X}$ by sampling from the latent space $\mathcal{Z}$ according to the modeled estimated density as in~\cite{kingma2018glow, tomINN}.
To ensure bijectivity and bidirectional execution, NFs are chains of invertible transformations.
Different architectures of NFs have been presented~\cite{realnvp, kingma, germain, maf}, which differ in the efficiency of their forward or backward pass.
Our proposed MVT-Flow is based on the popular Real-NVP~\cite{realnvp}, which efficiently computes both directions.

For anomaly detection with normalizing flows, density estimation is performed on normal data to calculate the likelihood of data points.
Anomalies are defined by having a small likelihood of being normal.
Numerous work is based on NFs for image-based anomaly detection~\cite{differnet, csflow, cflow, yan2022cainnflow, ast} by applying Real-NVP on pre-extracted feature maps.
However, this approach is not directly applicable to multivariate time series as the data domain is different which is crucial for the design of the NF.
Other work show applications on tabular data~\cite{nf_deep} and trajectories~\cite{nf_trajectory}.
Schmidt and Simic~\cite{nf_time_series} apply two variants of NFs, masked autoregressive flows~\cite{maf} and FFJORD~\cite{ffjord}, to industrial time series, but only evaluated on a single dataset with the motor current as the only signal.
Their approach is not applicable to our dataset, since the fully connected networks do not scale to the multitude of signals in \datasetname{}.
{To extend the applicability to multidimensional signals, Dai and Chen~\cite{GANF} propose to interpret the signals as a Bayesian network and use statistical dependencies as a condition for individual flows, each processing one signal.
Using the Bayesian network, which is estimated in advance, the individual likelihoods can then be combined.
To the best of our knowledge, we are the first to process multivariate time series using a single NF for AD.}

\section{Dataset}
Our dataset \datasetname{} contains machine data of a collaborative robot, which moves a can by performing an industrial pick-and-place task that is described in Sec.~\ref{pickplace}.
The samples $X=\{x_1, x_2, ..., x_{n_{rec}}\}$ consist of time series of machine data $x_i \in \mathbb{R}^{T \times S}$ with $T$ as the number of time steps, $S$ as the number of signals {and $n_{rec}=2122$ as the number of time series}, each recorded over one pick-and-place operation.
The data is split into a training set $X_{\mathrm{tr}}$ and a test set $X_{\mathrm{te}}$.
As usual in anomaly detection, the training set contains only normal data, which includes regular samples without anomalies.
The test set contains both, normal data and anomalies, including 12 diverse anomaly types as explained in Section~\ref{anomalies}.
In order to create a realistic scenario, we have divided the normal data into training and test data as follows:
Up to a certain period of time, only training data including 948 samples was recorded.
Subsequently, recordings of anomalies (755 samples) and normal data (419 samples) for the test set were taken alternately.
This simulates a real application where training data would be recorded first in the same way to train the model before the test case occurs.
{To exclude temperature effects, we let robots warm up for half an hour before each recording.}

\begin{figure}
  \centering
  \newcommand\PickImgWidth{0.2}
  \subfloat[][]{\includegraphics[width=\PickImgWidth\textwidth]{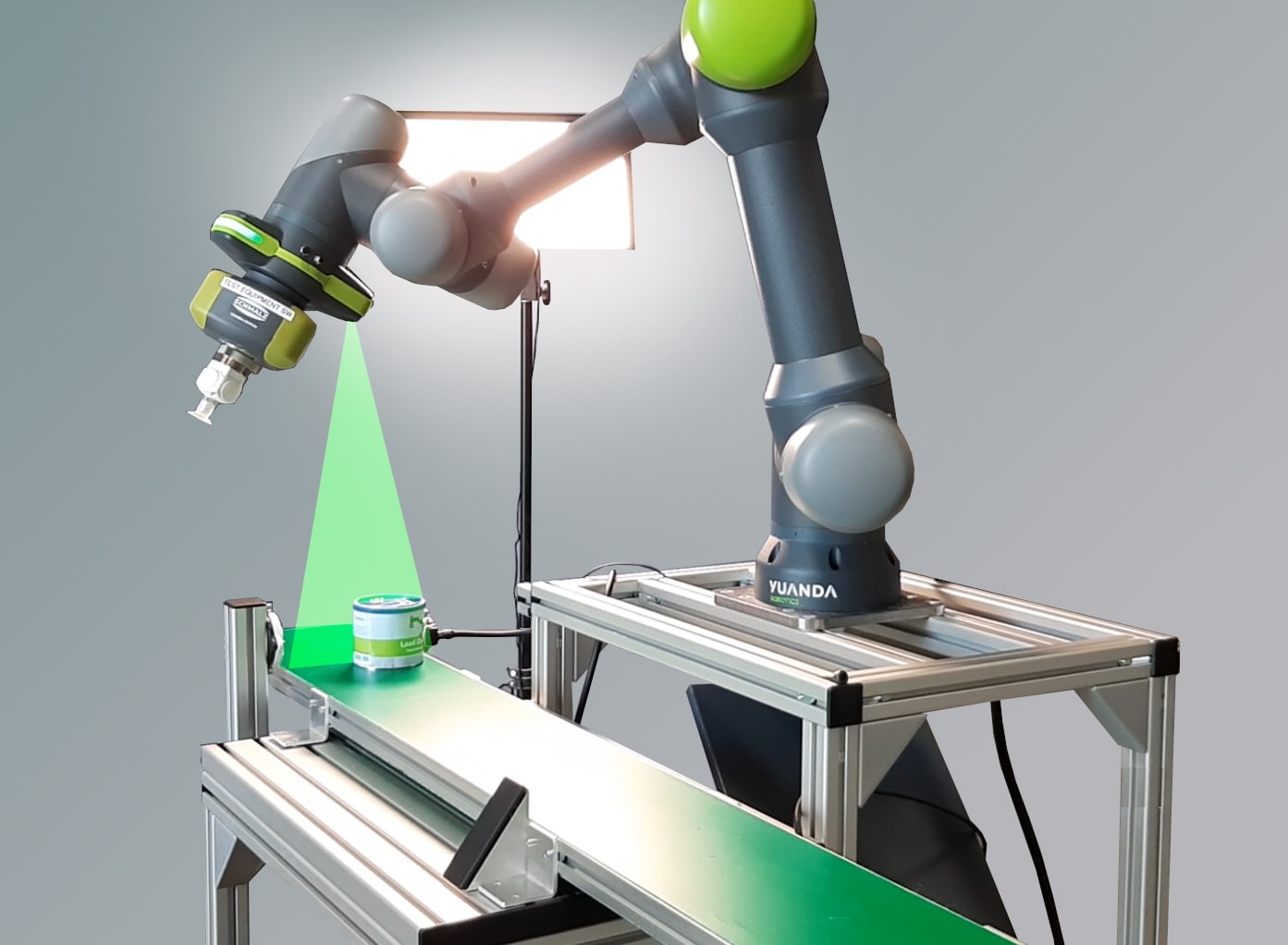}}%
  \subfloat[][]{\includegraphics[width=0.08\textwidth]{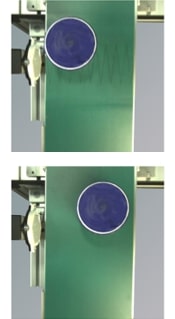}}%
  \subfloat[][]{\includegraphics[width=\PickImgWidth\textwidth]{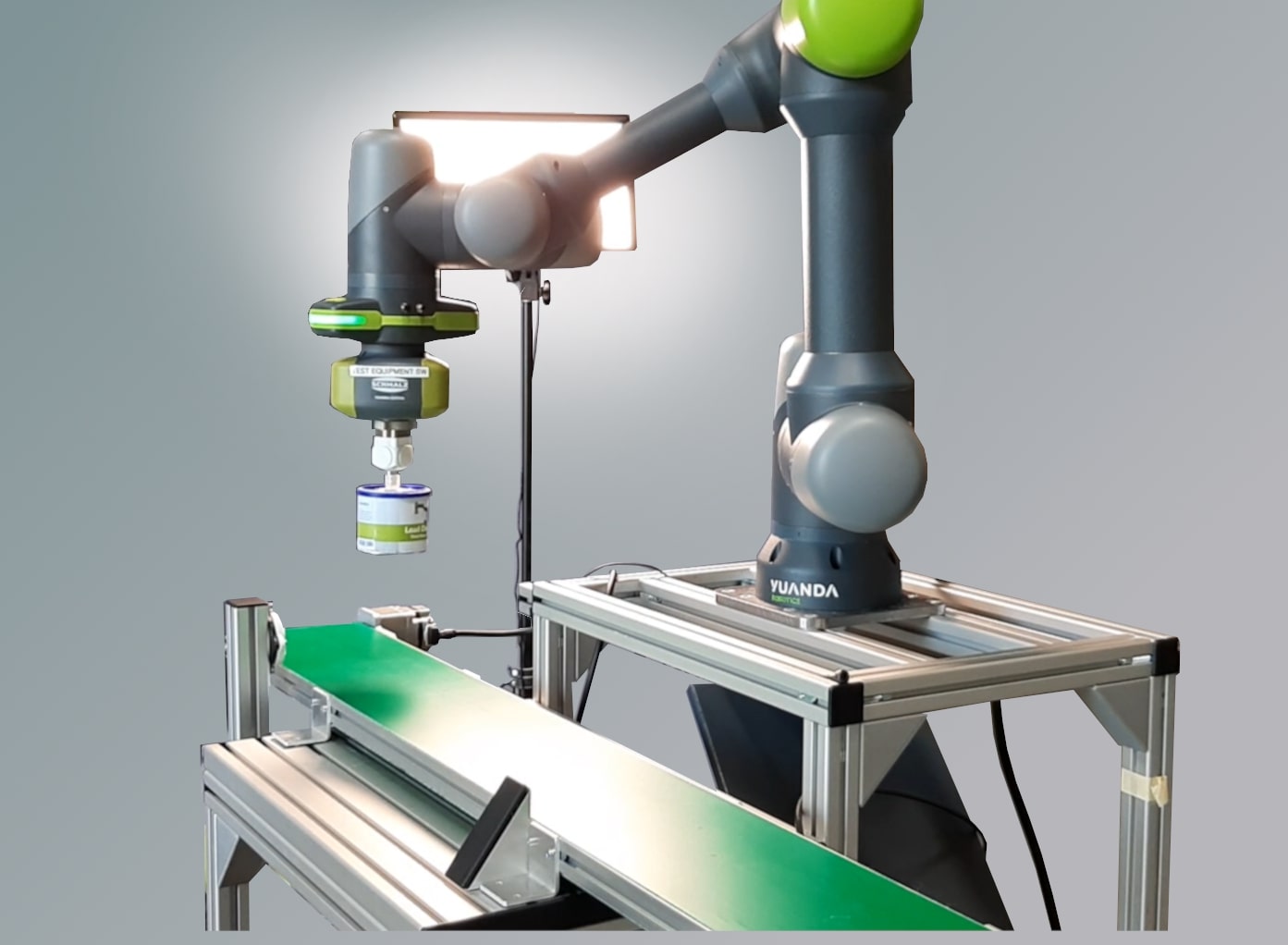}}%
  \caption{Scanning (a) the conveyor with the robot's built-in camera drawn in green, detecting (b) and gripping (c) the can from different positions.}
  \label{fig:object_detection}
\end{figure}

\begin{table}
\centering
\resizebox{1\linewidth}{!}{
\begin{tabular}{cllc}
\hline
\rowcolor[HTML]{EEEEEE} 
\textbf{No.} & \textbf{Action} & \textbf{Step} & \textbf{Record} \\ \hline
1 & \begin{tabular}[c]{@{}l@{}}Can detection\\ (camera-aided)\end{tabular} & \begin{tabular}[c]{@{}l@{}}1.1 Move to scan position\\ 1.2 Detect can position\end{tabular} & \checkmark \\ \hline
2 & Grip can & \begin{tabular}[c]{@{}l@{}}2.1 Move down\\ 2.2 Close gripper
\\ 2.3 Move up\end{tabular} & \checkmark \\ \hline
3 & Place can & \begin{tabular}[c]{@{}l@{}}3.1 Move to target position\\ 3.2 Move down\\ 3.3 Open gripper\\ 3.4 Move up to init. position\end{tabular} & \checkmark \\ \hline
4 & \begin{tabular}[c]{@{}l@{}}Create \\ variance \\ by random \\ placement\end{tabular} & \begin{tabular}[c]{@{}l@{}}4.1 Move down\\ 4.2 Close gripper\\ 4.3 Move up\\ 4.4 Move to random position\\ 4.5 Move down\\4.6 Open gripper\\4.7 Move up to init. position\end{tabular} & \xmark \\ \hline
\end{tabular}}
\caption{Actions of the pick-and-place application.
{The column \textit{Record} indicates if an actions is part of the dataset.}
}
\label{tab:steps}
\medskip
\footnotesize
\end{table}

\begin{figure*}[b]
  \centering
  
  \hspace{\imgSpaceBetween}
  \subfloat[][]{\includegraphics[width=\anoImgWidth\textwidth]{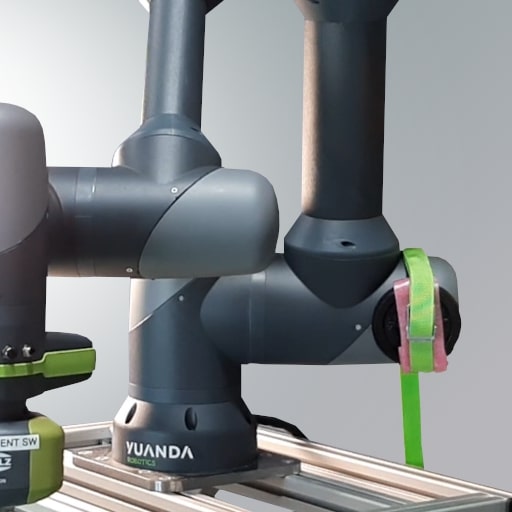}\label{fig:img_axis_weight}}%
  \hspace{\imgSpaceBetween}
  \subfloat[][]{\includegraphics[width=\anoImgWidth\textwidth]{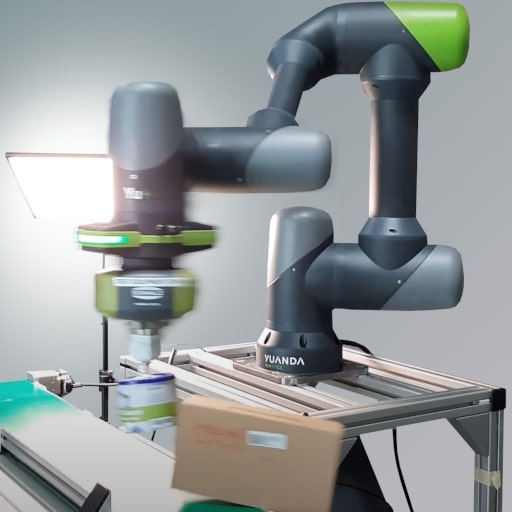}\label{fig:img_collision_package}}%
  \hspace{\imgSpaceBetween}
  \subfloat[][]{\includegraphics[width=\anoImgWidth\textwidth]{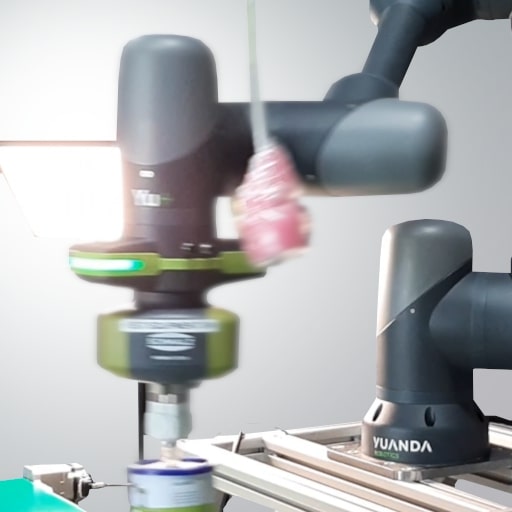}\label{fig:img_collision_cable}}%
  \hspace{\imgSpaceBetween}
  \subfloat[][]{\includegraphics[width=\anoImgWidth\textwidth]{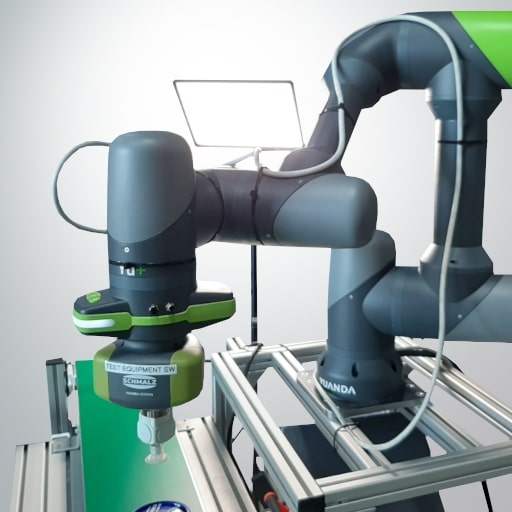}\label{fig:img_hanging_cable}}%
  \hspace{\imgSpaceBetween}  
  \subfloat[][]{\includegraphics[width=\anoImgWidth\textwidth]{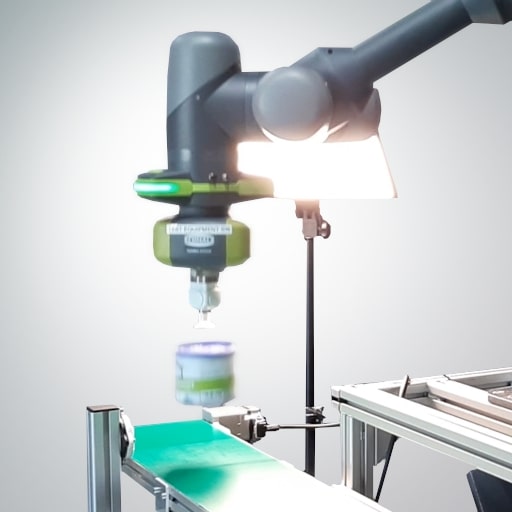}\label{fig:img_lose_can}}%
  \hspace{\imgSpaceBetween}  
  \subfloat[][]{\includegraphics[width=\anoImgWidth\textwidth]{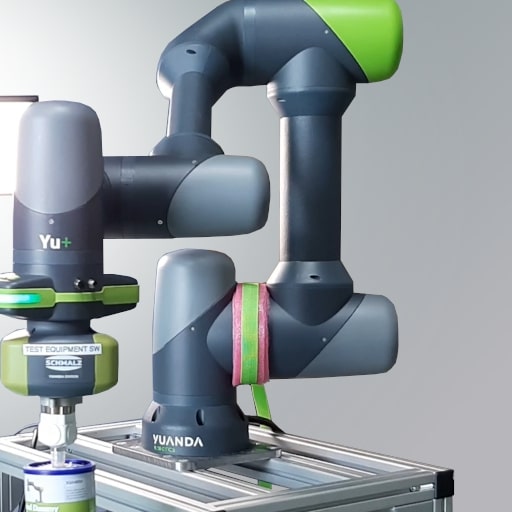}\label{fig:img_friction}}%
  
  \caption{Examples of induced anomalies: Increased moment of inertia caused by additional axis weight (a), collisions with boxes (b) and a hanging cable (c), cable routed at robot (d), losing the can during movement (e) and increased friction of an axis (f).}
  \label{fig:ano_examples}
    \vspace{-2mm}
\end{figure*}

\subsection{Setup}
The setup consists of a conveyor equipped with a light barrier, a robotic arm with vacuum gripper and the can to be moved as shown in Figure~\ref{fig:setup}.
We use the collaborative robotic arm~\textit{Yu-Cobot} with 6 axes {and an integrated camera}, which is mounted on a platform made of aluminum profiles.
Our robotic arm is equipped with a vacuum gripper {\textit{(Schmalz vacuum generator)}} for gripping the can.
The conveyor is also mounted on the platform in the working space of the robotic arm.
The conveyor control is connected to the digital outputs of the robotic arm so that the conveyor can be triggered by our robot control software.
To detect the presence of a can at the end of the conveyor belt, the light barrier connected to the digital inputs of the robotic arm can be used.
A light source permanently ensures that the internal camera of the robotic arm can also be used for visual object detection at night.

The machine data, which is further described in Sec.~\ref{robot_signals}, is transmitted between the robot controller and the individual axes via a field bus as a network interface (EtherCAT).
The software \textit{tshark5}\footnote{\href{
https://www.wireshark.org/docs/man-pages/tshark.html}{
https://www.wireshark.org/docs/man-pages/tshark.html}} is used to record all the data of this network interface.
\newcommand\aux{auxiliary}
\newcommand\nom{target}
\newcommand\meas{measurement}
\newcommand\ang{$\mathrm{rad}$}
\newcommand\trq{$\mathrm{Nm}$}
\newcommand\volt{$\mathrm{V}$}
\newcommand\pwr{$\mathrm{W}$}
\newcommand\cur{$\mathrm{A}$}
\newcommand\estimated{estimation}
\begin{table}[!t]
\centering
\resizebox{1\linewidth}{!}{
\begin{tabular}{llclll}
\hline

\rowcolor[HTML]{EEEEEE}  & \textbf{Feature} & $\bm{\forall}$\textbf{Axes} & \textbf{Signal type} & {\textbf{unit}} & \textbf{type} \\ \hline
\multirow{7}{*}{\rotatebox[origin=c]{90}{metadata}}
& time & \xmark & \aux & $\mathrm{s}$ & float \\ \cline{2-6}
& sample & \xmark & \aux & - & int \\ \cline{2-6}
& anomaly & \xmark & \aux & - & bool \\ \cline{2-6}
& category & \xmark & \aux & - & int \\ \cline{2-6}
& setting & \xmark & \aux & - & int \\ \cline{2-6}
& active & \xmark & \aux & - & bool \\ \cline{2-6}
& variant & \xmark & \aux & - & int \\ 
\hline
\multirow{15}{*}{\rotatebox[origin=c]{90}{mechanical}}
& target\_position & \checkmark & \nom & \ang & float \\ \cline{2-6} 
& target\_velocity & \checkmark & \nom& \ang$\mathrm{/s}$  & float\\ \cline{2-6}
& target\_accel. & \checkmark & \nom & \ang$\mathrm{/s²}$ & float \\ \cline{2-6}
& target\_torque & \checkmark & \nom & \trq  & float\\ \cline{2-6}
& computed\_inertia & \checkmark & \estimated & $\mathrm{kg \cdot m^2}$ & float \\ \cline{2-6}
& computed\_torque & \checkmark & \estimated & 
\trq & float \\ \cline{2-6}
& motor\_position & \checkmark & \meas & \ang  & float\\ \cline{2-6}
& motor\_velocity & \checkmark & \meas & \ang/s  & float\\ \cline{2-6}
& joint\_position & \checkmark & \meas & \ang  & float\\ \cline{2-6}
& joint\_velocity & \checkmark & \meas &  \ang/s  & float\\ \cline{2-6}
& motor\_torque & \checkmark & \estimated & \trq  & float\\ \cline{2-6}
& torque\_sensor\_a & \checkmark & \meas & \trq  & float\\ \cline{2-6}
& torque\_sensor\_b & \checkmark & \meas & \trq  & float\\ \cline{2-6}
& power\_motor\_mech  & \checkmark & \estimated & \pwr  & float\\ \cline{2-6}
& power\_load\_mech  & \checkmark & \estimated  & \pwr  & float\\
\hline
\multirow{10}{*}{\rotatebox[origin=c]{90}{electrical}}
& motor\_iq & \checkmark & \estimated & \cur  & float\\ \cline{2-6} 
& motor\_id  & \checkmark & \estimated & \cur  & float\\ \cline{2-6} 
& power\_motor\_el  & \checkmark & \estimated & \pwr  & float\\ \cline{2-6}
& motor\_voltage & \checkmark & \meas & \volt  & float\\ \cline{2-6}
& supply\_voltage & \checkmark & \meas & \volt  & float\\ \cline{2-6}
& brake\_voltage & \checkmark & \meas & \volt  & float\\ \cline{2-6}
& robot\_voltage  & \xmark & \meas & \volt  & float\\ \cline{2-6}
& robot\_current  & \xmark & \meas & \cur  & float\\ \cline{2-6} 
& io\_current  & \xmark & \meas & \cur  & float\\ \cline{2-6} 
& system\_current  & \xmark & \meas & \cur  & float\\ 
\hline
\end{tabular}
}
\caption{List of signals from our dataset, grouped according to metadata, mechanical and electrical signals. \textit{$\bm{\forall}$Axis} denotes an axis-specific signal which is given for all 6 axes.
{The metadata is further described in Sec.~\ref{robot_signals}.}}
\label{tab:signals}
    \vspace{-5mm}
\end{table}
\subsection{Robot and Signals}
\label{robot_signals}
We recorded the machine data of the robotic arm including all the 6 axes at a frequency of 500 Hz.
The machine data includes 130 signals in total which are listed in Table~\ref{tab:signals} and can be divided into target, measured, estimated and auxiliary signals.
The target signals are the states commanded by the robot controller, the measured signals comprise directly measured sensor signals from which further signals are estimated.
The auxiliary values are metadata which have been added subsequently for investigations of the dataset with the pick-and-place application and may not be entirely available in real operations.
They must not be used for machine learning approaches.
Measured signals related to each axis include joint positions, velocities, torques (2 sensors per axis), torque-forming current $I_{q}$ {and magnetizing current $I_q$}.
The absolute encoder \textit{RLS AksIM-2} and a \textit{Sensodrive GMS} were used for the position and torque measurements.
Equivalent to the measured values, the target signals for each axis contain the target position, the target velocity and the expected torque.
In addition to the 21 signals which are given for all 6 axes, 9 general electrical signals such as the supply voltage and the total current of the robotic arm, for example, are also part of our dataset.

The metadata includes the following information:
\begin{packed_enum}
    \item \textit{time}: elapsed time (in seconds) within the sample
    \item \textit{sample}: ID of the sample the data point is related to
    \item \textit{anomaly}: denotes if related sample contains an anomaly
    \item \textit{category}: denotes the anomaly type (see Table~\ref{tab:anolist})
    \item \textit{variant}: variant within the anomaly type (see Table~\ref{tab:anolist})
    \item \textit{active}: denotes if robot is currently moving
    \item \textit{action}: action ID within the sample (see Table~\ref{tab:steps})
\end{packed_enum}

\subsection{Pick-And-Place Operation}
\label{pickplace}
Our dataset was recorded by controlling the robot via a Python application that includes the algorithm of the pick-and-place cycles.
A recording cycle of the dataset proceeds as follows: The robot is moved to a fixed scanning position and first detects the position of a randomly placed can on the conveyor belt through its integrated camera and grips it (see Figure ~\ref{fig:object_detection}).
The robotic arm then moves the can to a fixed target position on the opposite side of the conveyor belt ending the recording cycle.
Finally, the can is placed at a random position and then conveyed back to the starting area for the next recording cycle which is triggered by the light barrier.

The {uniformly} random placement of the can introduces variance into the set of normal data, which is to be expected in this form in real applications.
Note that the random placement itself is not part of the recording since it only serves as an auxiliary task for the dataset design.
The steps of the entire recording process are described in more detail in Table~\ref{tab:steps} {and shown in our \href{
https://www.tnt.uni-hannover.de/vorausADrecording}{video}}\footnote{\href{
https://www.tnt.uni-hannover.de/vorausADrecording}{
https://www.tnt.uni-hannover.de/vorausADrecording}}.
As usual, most of the motions in path planning were joint angle optimized for time efficiency.
Only the \textit{move up} and \textit{move down} commands were performed with linear motion to ensure smooth picking and placing of the can.

\begin{table}
\centering
\resizebox{1\linewidth}{!}{
\begin{tabular}{llrl}
\hline
\rowcolor[HTML]{EEEEEE}  \textbf{Anomaly type} & \textbf{Cause} & {\textbf{No.}} & \textbf{Variants}\\ \hline
Additional friction & axis wear & 144 & 2 levels $\times$ 6 axes\\ \hline
Miscommutation & axis wear & 89 & 6 axes \\ \hline
Misgrip of can & gripping err. & 11 & 1 \\ \hline
Losing the can & gripping err. & 74 & random position \\ \hline
Add. axis weight & process err. & 156 & 3 weights $\times$ 6 axes\\ \hline
Collision w/ foam & process err. & 72 & 2 sizes $\times$ rand. pos. \\ \hline
Collision w/ cables & process err. & 48 & 2 types $\times$ rand. pos.\\ \hline
Collision w/ cardboard & process err. & 22 & random position\\ \hline 
Varying can weight & process err. & 80 & 6 weights \\ \hline
Cable routed at robot & process err. & 10 &  1 \\ \hline
Invalid gripping pos. & process err. & 12 & random position \\ \hline
Unstable platform & process err. & 37 & 3 levels \\ \hline
\end{tabular}}
\caption{List of anomaly types grouped in their cause. The listed variants are labeled in the dataset.}
\label{tab:anolist}
    \vspace{-5mm}
\end{table}

\subsection{Anomalies}
\label{anomalies}
For the test set, we intentionally induced anomalies of different types, which cover most cases of anomalies occurring in reality.
{In the selection of anomalies, we have created scenarios that represent errors in all actions (object detection, gripping, transport) and components (can, robot, platform) as well as the environment (collisions, objects on the robot) and further added those that we know from practical experience.}
We divide the anomalies into process errors, gripping errors, and robot axis wear.
An overview of all induced anomalies is given in Table~\ref{tab:anolist}.

\subsubsection{Process Errors} 
Process errors concern errors of the environment around the robot, which arise from a faulty setup.
Our dataset includes the following scenarios:
\begin{packed_enum}
    \item Additional axis weight:
    This might be caused by components that have not been disassembled from the robot.
    A weight of {115 g, 231 g or 500 g} was attached to each of the 6 axes as shown in Figure~\ref{fig:img_axis_weight}.
    This was done for each axis.
    The additional weight will affect all joint torques due to increased inertia and gravity. 
    \item Collisions:
    Collisions of the robot with objects can occur due to process flow errors or human errors. Our dataset includes collisions with lightweight objects such as foam, hanging cables, or empty cartons induced, as these are difficult to detect.
    The cartons and various foam cubes are placed at different positions so that recordings of collisions exist at different times.
    In some cases the objects are hit head-on, in others they are merely touched. Note that collisions sometimes occur even when the robot is not moving, since the previously hit object may swing back.
    Figure~\ref{fig:ano_examples} shows exemplary collisions with a carton~(b) and a hanging cable~(c).
    \item Varying can weight:
    As an example of disturbances in the process, runs with different weights of the can are included as anomalies.
    This may be the result of a damaged can or filling errors in a previous processing step.
    The can used in the normal data weighs {241 g} while recorded anomalies include 6 different weights between {41 g} (empty can) and {448 g}.
    \item Cable routed at robot:
    Cables or hoses are often routed along the robotic arm for various tools such as screwdrivers or painting equipment.
    If this cable is not dismantled again for another application, this can affect the current application.
    Included in this dataset are recordings in which a cable is attached to the robot with cable ties as shown in Figure~\ref{fig:img_hanging_cable}.
    {The cable causes friction on the axes when bending or twisting and due to gravity.}
    \item Invalid gripping position:
    An error in the process sequence, such as a worn conveyor belt with extended stopping distance, may result in unusual gripping positions of the can.
    Camera-based object detection can compensate for these errors as long as the can remains in the camera's field of view.
    This unusual gripping position nevertheless indicates an unusual process sequence.
    A time delay is integrated in the software of the robot application for recording unusual gripping positions.
    \item Unstable platform:
    In collaborative applications, robotic arms are often mounted on mobile platforms.
    An incorrectly adjusted mobile platform as well as uneven ground can lead to unwanted vibrations and thus affect the robotic arm.
    {We released the brakes under the platform to free the spring at the wheel and placed a different amount of foam under it.}
\end{packed_enum}

\subsubsection{Gripping Errors} 
Defects or contamination of the vacuum gripper may cause the can to become detached due to acceleration or centrifugal force. The loss of the can also happen during braking before the target position, which is why there are also samples in which the robot loses the can shortly before placement. To simulate the anomalies \textit{Losing the can} and \textit{Misgrip of can}, the gripper is timed to open during different movements of the robotic arm as shown in Figure~\ref{fig:img_lose_can} or the can is not gripped at all because the gripper is intentionally switched off. 

\subsubsection{Robot axis wear}
{The simulation of robot axis wear for the data set is a major challenge, as the wear must be induced in a limited period of time and at the same time not cause permanent damage.
In the datasets \cite{swvae, rnn_vae, zhang2021robot, aursad} no wear anomalies were considered at all, while \cite{hornung2014model} changed the maximum velocity of the robot axes.
However, wear of the gears affects the friction of the axis.
The motor control would adjust the motor current and thus the motor torque to maintain the specified path and, through this, the velocity. 
The wear of electrical and magnetic components would affect the efficiency and motor current of the robot axis instead of the velocity, without leading to increased motor torque.
The two cases have been simulated as separate anomaly types:}

\begin{packed_enum}
    \item Miscommutation:
    {The commutation of the motor is rotated with parameters in the firmware, i.e. deliberately calibrated incorrectly. This shifts the current curve of each phase to the respective optimal motor position. The proportion of torque-forming energy and thus the efficiency of the motor are reduced. This simulation has no effect on the path or velocity of the robot, as would be the case if the motor components were subject to mechanical and magnetic wear.}
    For each case the recordings contain a single incorrectly commuted axis.
    \item Increased friction:
    Dirt or age-related wear can lead to increased friction of individual robot axes. 
    {The friction can occur at the joint transitions or in the gearbox of the robot axis, which is why the friction was also induced on the output side for our dataset.}
    The additional friction is generated by attaching foam at the transition point of two robot axes.
    The foam is fastened with a tension belt as shown in Figure \ref{fig:img_friction}.
    {Due to the friction, a higher torque of the motor is needed for the same movement.}
    In each sample of this anomaly type, there is one axis with increased friction in one of two strength levels. 
\end{packed_enum}

\begin{table}
\centering
\resizebox{1\linewidth}{!}{
    \begin{tabular}{lll}
    \hline
    \rowcolor[HTML]{EEEEEE}
    \textbf{Dataset} & \textbf{AURSAD}~\cite{aursad} & \textbf{\datasetname~(ours)} \\ \hline
    \textbf{Application} & \textbf{screwdriver} & \textbf{pick-and-place} \\ \hline
    \textbf{Variance (normal data)} & \textbf{discrete (screw holes)} & \textbf{continuous (rand. position)} \\ \hline
    Robot & UR3e & Yu-Cobot \\ \hline 
    \#Axes & 6 & 6 \\ \hline
    \textbf{Anomaly types} & \textbf{4} & \textbf{12} \\ \hline
    \#Normal samples & 1420 & 1367 \\ \hline
    \#Anom. samples & 625 & 755 \\ \hline
    \textbf{$\varnothing$sample duration} & \textbf{15.3s} & \textbf{11s} \\ \hline
    \textbf{Signal frequency} & \textbf{100 Hz} & \textbf{500 Hz} \\ \hline
    \#Signals & 125 & 130 \\ \hline
    \textbf{Raw dataset size} & \textbf{5.81 GB} & \textbf{11.17 GB} \\ \hline
    \end{tabular}}
\caption{Comparison between AURSAD and \datasetname. The main differences are highlighted in bold.}
\label{table:dataset_comparison}
    \vspace{-4mm}
\end{table}

\subsection{Comparison to AURSAD}
\label{dataset_comparison}
Table~\ref{table:dataset_comparison} compares the key characteristics from our dataset with AURSAD~\cite{aursad}, which is currently the only comparable dataset in this domain.
The different task of the robot (screwdriver vs. pick-and-place in our case) results in different challenges for the detection:
The variance of the normal data in AURSAD is nearly discrete, as the screws are driven into pre-defined holes, whereas the samples of our dataset are continuously distributed, since the can is at a random, not pre-defined location on the conveyor belt before gripping it.
While AURSAD contains only 4 anomaly types, which are exclusively related to the screw and plate, \datasetname{} contains 12 diverse anomaly types, which cover most possible scenarios in reality.
Most of our anomaly types (7), such as collisions or axis-wear-related anomalies, do not only occur exclusively in pick-and-place tasks, but are general potential faults in robotics applications.
Thus, evaluations on our dataset provide insights for other robotics applications as well.
The machine data of \datasetname{} was provided with a significantly higher sampling rate ($500Hz$ vs. $100Hz$), which allows the detection of high-frequency deviations.
In contrast to AURSAD, our dataset includes measurements of torque sensors instead of the force on the end effector.

\subsection{Analysis}
\begin{figure}[!t]
    \centering
    \includegraphics[width=1\linewidth]{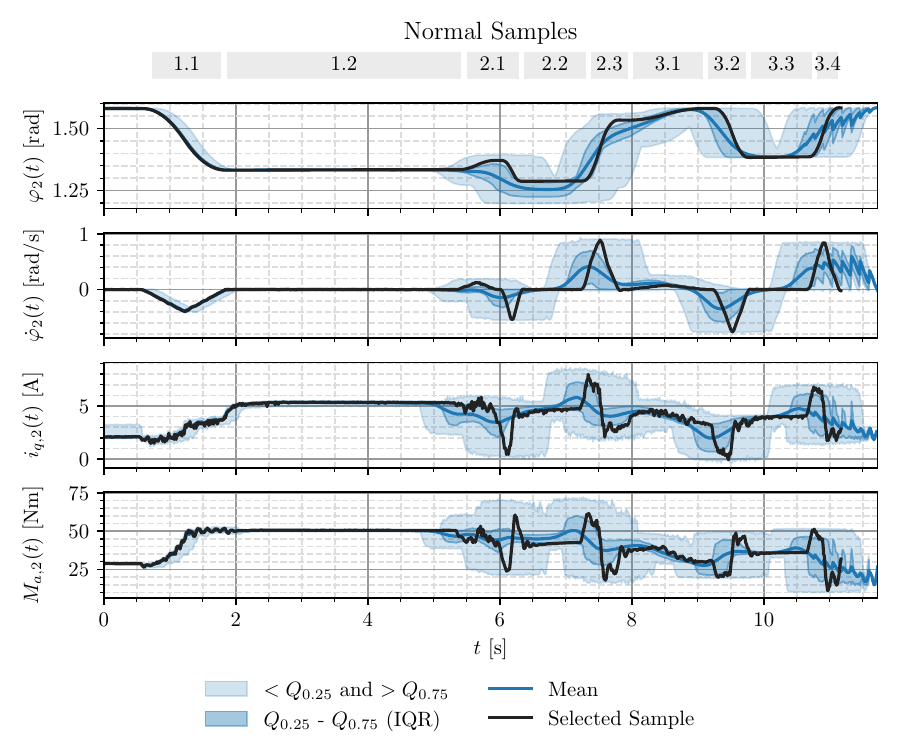}
    \caption{Selected signals from the second axis of the normal data within our \datasetname{} dataset. The light blue areas represent the different quartils, while the blue line shows the mean signal and the black line a random sample. At the top, the individual actions from Table~\ref{tab:steps} are shown.}
    \label{fig:normal_quali}
    \vspace{-4mm}
\end{figure}

\begin{figure}[!htb]
    \centering
    \subfloat[Collision with foam]{\includegraphics[width=1\linewidth]{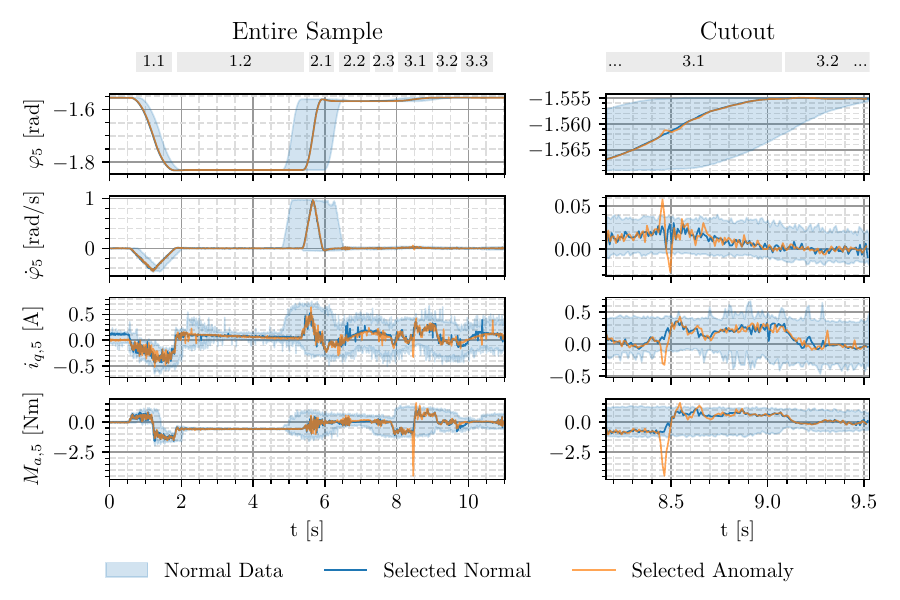} \label{fig:ano_collision}}\\
    \vspace{-2mm}
    \subfloat[Losing the can]{\includegraphics[width=1\linewidth]{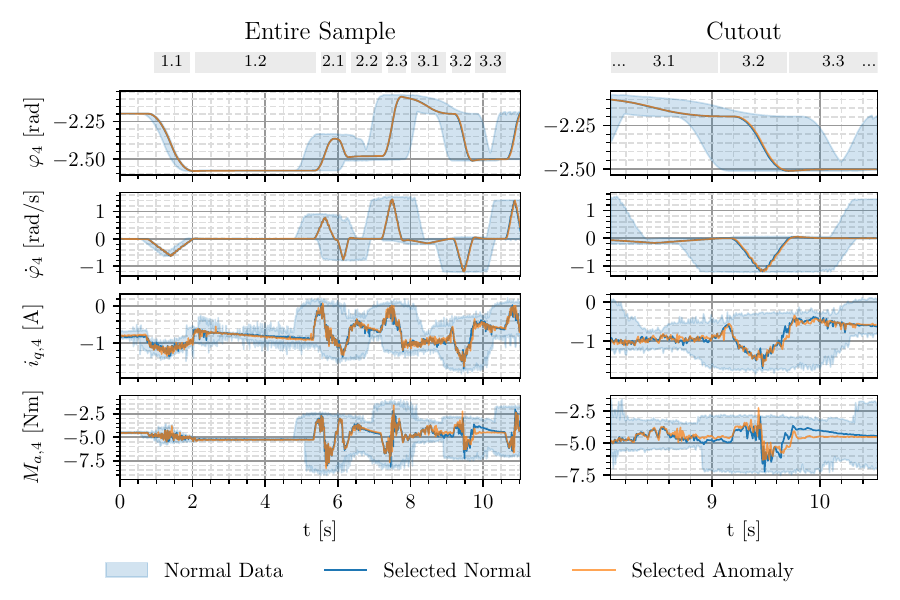} \label{fig:ano_losing}}
    \caption{Selected signals of the anomalies \textit{collision with foam} and \textit{losing the can}. The left plot shows the entire sample, while the right side zooms into the time period of the anomalous event. For the normal sample, we visualize the most similar example in the dataset.}
    \label{fig:ano_signals}
    \vspace{-3mm}
\end{figure}

\subsubsection{Normal Data}
Figure~\ref{fig:normal_quali} shows selected signals of the second robot axis from an exemplary recording from the normal data of the \datasetname{} dataset.
Between second zero and one, the robot moves to the fixed scan position and remains there to detect the can.
The minimum and maximum values are close to each other for the first 4.5 seconds since all movements up to this point are identical and the recording always starts at the same time.
Starting from the visual object detection of the can, the variance is increased due to the temporal and spatial offset caused by the movement to different positions of the can.
In the following actions {1-3} of Table~\ref{tab:steps} are performed as marked at the top of Figure~\ref{fig:normal_quali}.

\subsubsection{Anomalies}
Figure~\ref{fig:ano_collision} shows selected signals from an anomaly where the the robotic arm collides with a foam cuboid.
The collision occurs at about $t = 8.45$ s as visible on all signals by a deflection in the signal.
After the robotic arm has deviated from its joint angle target due to the collision the closed-loop control counteracts to correct the path which influences the joint angle $\varphi_5$ and its velocity $\dot{\varphi}_5$.
The deviation of the torque $M_{s1,5}$ is explained by its physical relationship with external forces acting on the robot.
\\The signals of the fourth axis from the anomaly \textit{losing the can} are shown in Figure~\ref{fig:ano_losing}.
An effect on the joint angle $\varphi_4$, the joint angular velocity $\dot{\varphi}_4$ and the motor current $i_{q_,4}$ is not visually apparent.
However, the torque $M_{s1,4}$ of the fourth axis shows a short oscillation at $t = 8.6 $ s, when the robotic arm loses the can.
After the loss of the can, for {$8.8\text{ s} < t < 9.6\text{ s}$}, a lower absolute torque is measured which can be explained by the decrease in weight as a result of the loss of the can.
In the time around {$t = 10\text{ s}$}, an increased absolute torque is measured compared to the error-free recording, since there is no contact between the conveyor belt and the (lost) can which would dampen the weight force.

\section{Method}
\label{nf_method}
In the following, we present a new baseline for anomaly detection on multivariate time series.
{Similar to~\cite{differnet, csflow, cflow}}, the method is based on density estimation of normal data via Normalizing Flows whose architecture we tailored to multivariate time series.
The estimated likelihoods of the NF are used as an indicator for anomalies:
We assume that anomalies have a low likelihood whereas the likelihood of normal data should be high.

Our MVT-Flow $f:\mathcal{X} \rightarrow \mathcal{Z}$ transforms the unknown data distribution $p_X$ of normal samples $x \in X_{\mathrm{tr}}$ to a target space $\mathcal{Z}$ with known {base} distribution $p_Z$.
Since $f$ is bijective, this enables us to measure likelihoods of data points after mapping them to the target space using $z = f(x)$.
Using a standard multivariate Gaussian distribution $\mathcal{N}(0,\,I_d)$ as target distribution with $d$ as the number of dimensions of $z$, the density of the target space is defined by
\begin{equation}
    p_Z(z) = \frac{1}{(2 \pi)^{d/2}} e^{-\frac{1}{2}\norm{z}_2^2}.
\label{eq:gaussian}
\end{equation}
Following the change of variables formula, the Jacobian of $f$, the likelihood of data points is given by 
\begin{equation}
\label{eqn:change_of_variables}
    p_X(x) = p_Z(z) \abs{
    \det{
    \frac{\partial z}
        {\partial x}
    }}
    .
\end{equation}
The final decision whether the sample is classified as an anomaly is given by thresholding the likelihood
\begin{equation}
    \mathcal{A}(x) = 
      \begin{cases}
        1 & \text{for } p_X(x) < \theta \\
        0 & else
      \end{cases}
\end{equation}
with $\theta$ as an adjustable parameter.
{In practice, the parameter can be set according to a previously defined acceptable false positive rate using the statistics of normal data.
In the following, we describe Real-NVP~\cite{realnvp} as the basis of our NF in Sec.~\ref{sec:foundations}, our extension MVT-Flow in Sec.~\ref{sec:mvtflow} and its training procedure in Sec.~\ref{training}.}

\subsection{Foundations}
\label{sec:foundations}
We first recap Real-NVP-based NFs~\cite{realnvp}.
{For more details, we refer the reader to introductory literature for Normalizing Flows~\cite{papamakarios2021normalizing,kobyzev2020normalizing, nf} and the original Real-NVP paper~\cite{realnvp}.}

A NF consists of a chain of \textit{coupling blocks} $$c_0 \circ c_1 \circ ... \circ c_{n_\mathrm{blocks}} $$ each of them performing an affine transformation.
Figure~\ref{fig:realnvp} shows the architecture of such a coupling block.
For simplicity of notation, the following explains the network with only a single coupling block.
First, the entries of the input $x$ are reordered by a pre-defined, usually randomly determined, permutation.
Consecutively, the signal is divided into the parts $x_1$ and $x_2$, from which {element-wise} scaling and translation coefficients for the counterpart are calculated from each other consecutively by applying the internal networks $g_{1}$ and  $g_{2}$.
{With this structure, correlations between $x_1$ and $x_2$ can be exploited by manipulating one component based on the other component until the desired base distribution is approximated. 
The shuffling permutation in each block enables that (after splitting) the correlations between the data components are exploited as much as possible.
The element-wise scaling and translation components are used to form a simple invertible mapping with an efficient calculation of the Jacobian (see Eq.~\ref{eq:loss}), which is needed for the likelihood calculation in Eq.~\ref{eq:loglikelihood}.
The internal networks can be realized by any differentiable neural network whose parameters are learned during the training as described in Section~\ref{training}.
Our design of $g_{1}$ and  $g_{2}$ is given in Sec.~\ref{sec:mvtflow}.}
The scaling and translation coefficients do have the same dimensions as the respective counterpart and are applied as following:

\begin{figure}[]
    \centering
    \includegraphics[width=1\linewidth, page=2]{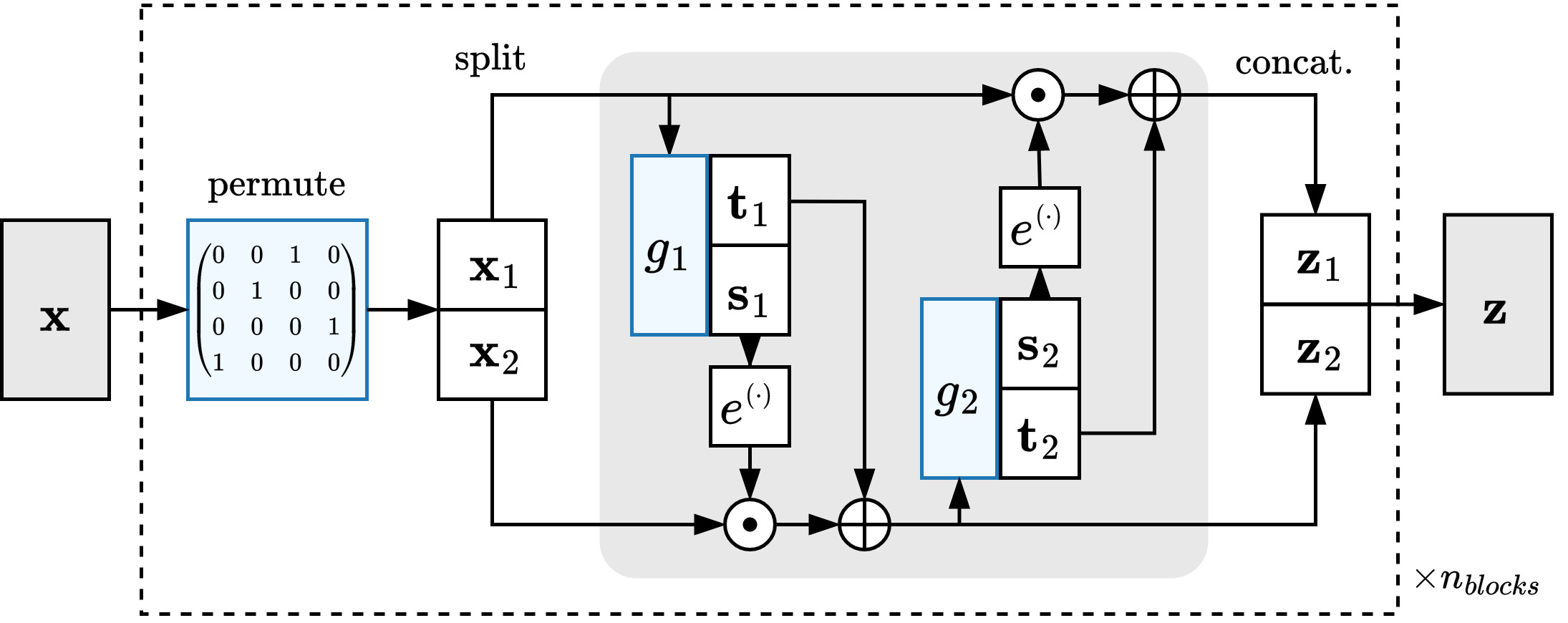}
    \caption{Architecture of a coupling block in Real-NVP.
    Symbols $\odot$ and $\oplus$ denote element-wise multiplication and addition.
    The design of the blue parts is domain-specific and our contribution in MVT-Flow.}
    \label{fig:realnvp}
    \vspace{-2mm}
\end{figure}

\begin{equation}
 \begin{aligned}
y_{2} = x_2 \odot e^{s_1} + t_1 \quad \text{with}\quad [s_1, t_1] = g_1(x_1)\\
y_{1} = x_1 \odot e^{s_2} + t_2\quad \text{with}\quad [s_2, t_2] = g_2(y_2)
\label{eq:coupling}
\end{aligned}
\end{equation}
with $\odot$ as the Hadamard product.
Finally, the output $y$ is the concatenation of $y_1$ and $y_2$ with the same number of dimensions as $x$.
{Since the permutation, addition, and multiplication operations are invertible, the overall network is invertible and therefore forms a diffeomorphism which enables a density estimation.}
The exponentiation of the scaling coefficients prevents the scaling coefficient from being zero to maintain invertibility.
A challenge in learning the density lies in the architecture design of the internal networks in combination with the permutation and split strategy.

\subsection{MVT-Flow}
\label{sec:mvtflow}
We adapt the design of the normalizing flow to exploit the structure of multivariate time series to appropriately model the data distribution {by defining the permutation strategy as well as the design of the internal networks $g_1$ and $g_2$.}
The time series have dimensions $T \times S$ with $T$ as the number of time steps and $S$ as the number of signals.
We exploit and maintain the temporal structure by using convolutions for the internal functions instead of fully-connected blocks as in~\cite{nf_time_series}.
{Thereby, we follow~\cite{csflow, ast}, in which state-of-the-art performance for anomaly detection on image data with convolutions in the internal networks of NFs was shown.}
The partitioning into signals is maintained as well.
However, the signals not longer have their original meaning during the transformation steps and should be interpreted as latent signals at the output, which are independent of each other.

\begin{figure}[]
    \centering
    \includegraphics[width=1\linewidth]{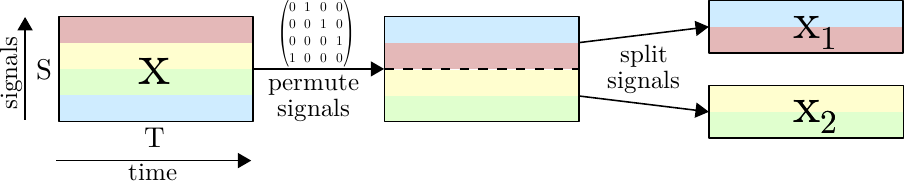}
    \caption{Permutation and split procedure inside a coupling block of MVT-Flow: First, the signals are permuted in a fixed, randomly defined, manner. After that, the sample is divided into two components, each comprising half of the signals. Best viewed in color.}
    \label{fig:nf_split}
    \vspace{-2mm}
\end{figure}

A coupling block of our MVT-Flow is designed as follows:
First, the signals of the input are reordered using a permutation matrix $P\in \{0,1\}^{S\times S}$ which is randomly determined initially for each block.
We split the sample along the signal axis evenly into the parts $x_1$ and $x_2$ with dimensions $T\times \frac{S}{2}$.
This enables the usage of correlations between different signals for density estimation.
Figure~\ref{fig:nf_split} visualizes the permutation and split procedure.

\begin{figure}[]
    \centering
    \includegraphics[width=1\linewidth]{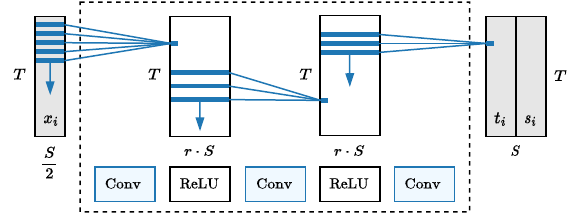}
    \caption{Architecture of the internal networks in MVT-Flow.}
    \label{fig:internal}
    \vspace{-2mm}
\end{figure}

The parts are then fed into the internal networks $g_1$ and $g_2$ to transform the counterpart subsequently according to Figure~\ref{fig:realnvp} and Eq.~\ref{eq:coupling}.
The internal networks consist of a sequence of 3 one-dimensional convolutions, whose kernel is moved along the time, and ReLUs as nonlinearities as shown in Figure~\ref{fig:internal}.
After the last convolutional layer, the output is split evenly along the signal axis to obtain the scaling and translation coefficients $s$ and $t$, {matching the dimensions $T\times \frac{S}{2}$ of $x_1$ and $x_2$.}
Since the exponentiation of the scaling coefficients (see Eq.~\ref{eq:coupling}) can cause an unstable optimization due to exploding gradients, we apply soft-clamping as proposed by~\cite{cinn} for better convergence:
\begin{equation}
\sigma_{\alpha}(s) = \frac{2\alpha}{\pi}\arctan{\frac{s}{\alpha}}
\end{equation}
with $\alpha$ as a hyperparameter that controls the clamping magnitude by restricting the values to the interval $(-\alpha, \alpha)$.

\subsection{Training}
\label{training}
During training, we optimize the parameters for the internal networks such that the training data in the latent space is normally distributed.
The overall model is trained via maximum-likelihood training:
The probability of the training data according to the model should be maximized.
Instead of maximizing the likelihood, we minimize the negative log-likelihood $- \log{p_X(x)}$ for convenience as it is equivalent and numerically more practicable.

The negative log likelihood of a standard multivariate Gaussian distribution is
\begin{equation}
-\log{p_Z(z)} = \frac{1}{2}\norm{z}_2^2 + C
 \label{eq:nllz}
\end{equation}
with a constant $C$ which is ignored for optimization.
Based on the log formulation of Equation~\ref{eqn:change_of_variables}
\begin{equation}
 \begin{aligned}
    - \log{p_X(x)} = - \log{p_Z(z)}  - \log{\abs{
    \det{
    \frac{\partial z}
        {\partial x}
    }}}
 \end{aligned}
 \label{eq:loglikelihood}
\end{equation}
we now insert Equation~\ref{eq:nllz} and use that the log determinant of the Jacobian is the sum of all scaling coefficients~\cite{realnvp}, which gives the final loss function
\begin{equation}
 \begin{aligned}
    \mathcal{L} = - \log{p_X(x)} = \frac{1}{2}\norm{z}_2^2    - \sum_{s \in \mathcal{S}} s
 \end{aligned}
 \label{eq:loss}
\end{equation}
with $\mathcal{S}$ as the set of all scaling coefficient entries.

\subsection{Temporal Analysis}
\label{temp_analysis}
In addition to the sequence-level detection which provides if any unusual event has occurred, in practice it is often relevant to analyze when this event happened.
This helps to find the cause of the error and to possibly eliminate it.
We provide a measure which indicates the impact of each time step on the anomaly score.
Similar to~\cite{differnet}, we use the gradient of the anomaly score with respect to the inputs by backpropagating the loss from Eq.~\ref{eq:loss}.
The input gradient $\nabla_x \in \mathcal{R}^{T \times S}$ is aggregated to a gradient for every time step $t$ by taking the $\ell_1$-norm over all signals:

\begin{equation}
    \nabla_x^t = \sum_{s=1}^{S}{\abs{\nabla_x(t, s)}}
    .
    \label{eq:gradient}
\end{equation}

\begin{table}[b]
\centering
\begin{tabular}{lc}
\hline
\rowcolor[HTML]{EEEEEE}
\textbf{General Parameters} & Value \\ \hline
Signal frequency & $100~Hz$  \\ \hline 
Batch size & 32  \\ \hline
Optimizer & Adam~\cite{adam} \\ \hline
Learning rate  & $8\cdot 10^{-4}$ \\ \hline 
Learning rate decay & 0.1 \\ \hline
Decay epochs &  11, 61 \\ \hline
Total epochs & 70 \\ \hline
\rowcolor[HTML]{EEEEEE}
\textbf{NF Parameters} &  \\ \hline
Coupling blocks $n_{\mathrm{blocks}}$ & 4 \\ \hline 
Signal scaling $r$ & 2 \\ \hline 
Kernel sizes $k_1, k_2, k_3$ & 13, 1, 1 \\ \hline 
Dilation $d_1, d_2, d_3$ & 2, 1, 1 \\ \hline 
\end{tabular}
\caption{Hyperparameter setting for MVT-Flow}
\label{table:hp}
    \vspace{-3.5mm}
\end{table}

\begin{table*}[h!b]
\begin{center}
\footnotesize
\begin{tabular}{l c c c c c c c} 
    \hline
     \rowcolor[HTML]{EEEEEE} 
    \textbf{Method} &
     \begin{tabular}{c} 1-NN \\ \cite{amer2012nearest}\end{tabular} & \begin{tabular}{c} PCA \\ \cite{hornung2014model}\end{tabular} & \begin{tabular}{c} CAE \\ adapts \cite{swvae}\end{tabular}  &\begin{tabular}{c} {LSTM-VAE} \\ \cite{park2018multimodal}\end{tabular} & \begin{tabular}{c} {GANF} \\ \cite{GANF}\end{tabular} & \begin{tabular}{c} {HMM} \\ \cite{hmm_ad}\end{tabular} & \begin{tabular}{c} MVT-Flow \\ \textbf{(ours)}\end{tabular} \\ \hline
\footnotesize
    Add. Friction & 74.8 & 76.4 & 89.4 $\pm$ 0.2  & 88.7 $\pm$ 1.5 & 88.5 $\pm$ 4.5& 88.0 $\pm$ 0.7 & \textbf{96.6} $\pm$ 0.6\\ \hline
    Miscommutation & 80.8 & 87.0 & 99.1 $\pm$ 0.0  & 98.1 $\pm$ 0.5 & 98.8 $\pm$ 1.1 & 93.7 $\pm$ 1.3 & \textbf{99.8} $\pm$ 0.3\\ \hline
    Misgrip can & \textbf{100.0} & \textbf{100.0} & \textbf{100.0} $\pm$ 0.0  & \textbf{100.0} $\pm$ 0.0 & 47.6 $\pm$ 13.1 & 71.0 $\pm$ 3.3&  95.3 $\pm$ 3.3\\ \hline
    Losing can  & 68.7 & 70.1 & 72.6 $\pm$ 0.3  &  70.4 $\pm$ 2.9 & 72.1 $\pm$ 5.8 & 88.8 $\pm$ 1.0&  \textbf{96.2} $\pm$ 0.4\\ \hline
    Add. axis weight & 75.0 & 79.2 & 93.5 $\pm$ 0.1 &82.7 $\pm$ 1.1  &93.2 $\pm$ 2.4 & 89.6 $\pm$ 1.2 & \textbf{94.1} $\pm$ 0.7\\ \hline
    Coll. foam & 69.6 & 73.9 & 81.5 $\pm$ 0.2  &  81.5 $\pm$ 2.1& 81.2 $\pm$ 6.2& \textbf{89.8} $\pm$ 1.5 &  87.5 $\pm$ 1.2\\ \hline
    Coll. cables & 74.5 & 75.7 & 79.6 $\pm$ 0.3 &  77.3 $\pm$ 3.6 & 82.7 $\pm$ 6.4 & \textbf{91.8} $\pm$ 1.4&  84.7 $\pm$ 1.2\\ \hline
    Coll. cardboard & 82.8 & 83.6 & 78.6 $\pm$ 0.3 &  82.8 $\pm$ 4.8  & 77.5 $\pm$ 5.8 & 86.3 $\pm$ 1.2&  \textbf{88.3} $\pm$ 1.2\\ \hline
    Var. can weight& 63.7 & 64.2 & 72.3 $\pm$ 0.3 &71.4 $\pm$ 2.1  & 68.9 $\pm$ 8.7 & \textbf{90.9} $\pm$ 2.0 &   85.1 $\pm$ 1.1\\ \hline
    Cable at robot & 63.0 & 71.6 & 83.1 $\pm$ 0.3 & 96.0 $\pm$ 1.4 & 76.6 $\pm$ 8.1 & 84.4 $\pm$ 1.1 &  \textbf{100.0} $\pm$ 0.0\\ \hline
    Invalid grip. pos. & 93.4 & 92.1 & 88.8 $\pm$ 0.3 & 97.7 $\pm$ 1.3 & 86.6 $\pm$ 10.8 & 91.8 $\pm$ 1.4 &  \textbf{100.0} $\pm$ 0.0\\ \hline
    Unstable platform & 83.6 & 85.9 & 83.9 $\pm$ 0.2  & 93.6 $\pm$ 1.9 & 84.6 $\pm$ 3.8 & 82.5 $\pm$ 1.2 &  \textbf{96.1} $\pm$ 0.7\\ \hline
  \rowcolor[HTML]{EEEEEE} \textbf{Mean} & 77.5 & 80.0 & 85.2 $\pm$ 9.2 & 86.7 $\pm$ 10.1  &79.9 $\pm$ 12.7 & 87.4 $\pm$ 5.8  & \textbf{93.6} $\pm$ 5.7\\  \hline
    \end{tabular}
\end{center}
\vspace{-0.15cm}
\caption{Anomaly detection results on \datasetname{} for MVT-Flow and other baselines measured in AUROC percentage. MVT-Flow shows the best performance for most categories and {on average}.}
\label{table:results}
\vspace{-0.35cm}
\end{table*}

\section{Experiments}
In this section, we first present the experimental setup along with evaluation metrics and implementation details (Sec.~\ref{metrics} and ~\ref{imp_details}).
We then compare our baseline MVT-Flow with previous work in Sec.~\ref{results}.
Finally, in Sec.~\ref{ablations} the characteristics of our dataset are varied to explore its impact on the detection performance.
\subsection{Metrics}
\label{metrics}
Typically, anomaly detection methods are evaluated using the \textit{receiving operator characteristic} (ROC) or the area under its curve (AUROC).
The ROC plots the relation between true positive rate (TPR) and false positive rate (FPR) for a varying threshold $\theta \in (-\infty, \infty)$ of a binary decision.
The AUROC has a range of $[0, 1]$:
A value of $1$ means that all samples can be classified correctly since the classes are perfectly separable.
Random decision making results in a value of $0.5$.

We decide for the AUROC as the main metric since it is independent of the number of anomalies, which is unknown and varies in practice, and independent of a particular threshold which is highly dependent on the application and environment.
The AUROC should be measured for each anomaly type individually to identify the strengths and weaknesses of a method.
For an overall comparison, it should be averaged over all 12 types.
We do not recommend calculating the area under one single ROC over all samples and anomaly types, as the number of samples per anomaly type strongly biases the metric in this case.

\subsection{Implementation Details}
\label{imp_details}

{In the following, we describe the implementation details of MVT-Flow and the other baselines we compare. As further baselines we use a PCA-based approach~\cite{hornung2014model}, 1-nearest-neighbor distance (1-NN~\cite{amer2012nearest}), a convolutional autoencoder (CAE), a LSTM-based VAE~\cite{park2018multimodal}, a Hidden Markov Model~\cite{hmm_ad} and a graph-augmented normalizing flow (GANF~\cite{GANF}).
We have used a sampling frequency of 100 Hz for all methods and provide a uniform dimensionality by padding samples to the maximum length for all experiments and all methods, unless otherwise stated.}
\subsubsection{MVT-Flow}
Table~\ref{table:hp} summarizes the hyperparameters of our method for the following experiments
We apply a learning rate decay during training to achieve a fast but stable convergence. The initial learning rate of $8\cdot 10^{-4}$ is multiplied by a factor of $0.1$ in epochs $11$ and $61$.
The training ends after a fixed number of 70 epochs.
For the NF, individual kernel sizes and dilation parameters for the 3 convolutional layers inside the internal networks are used.
While the first convolution takes 13 sample points with a dilation of 2, the other 2 layers process the signal locally having a kernel size of 1.
The number of signals is scaled in the hidden layers with a factor of $r=2$.
In total, the NF is a chain of 4 blocks, each using the same hyperparameters.

\begin{figure*}

    \centering
    \includegraphics[width=1\linewidth]{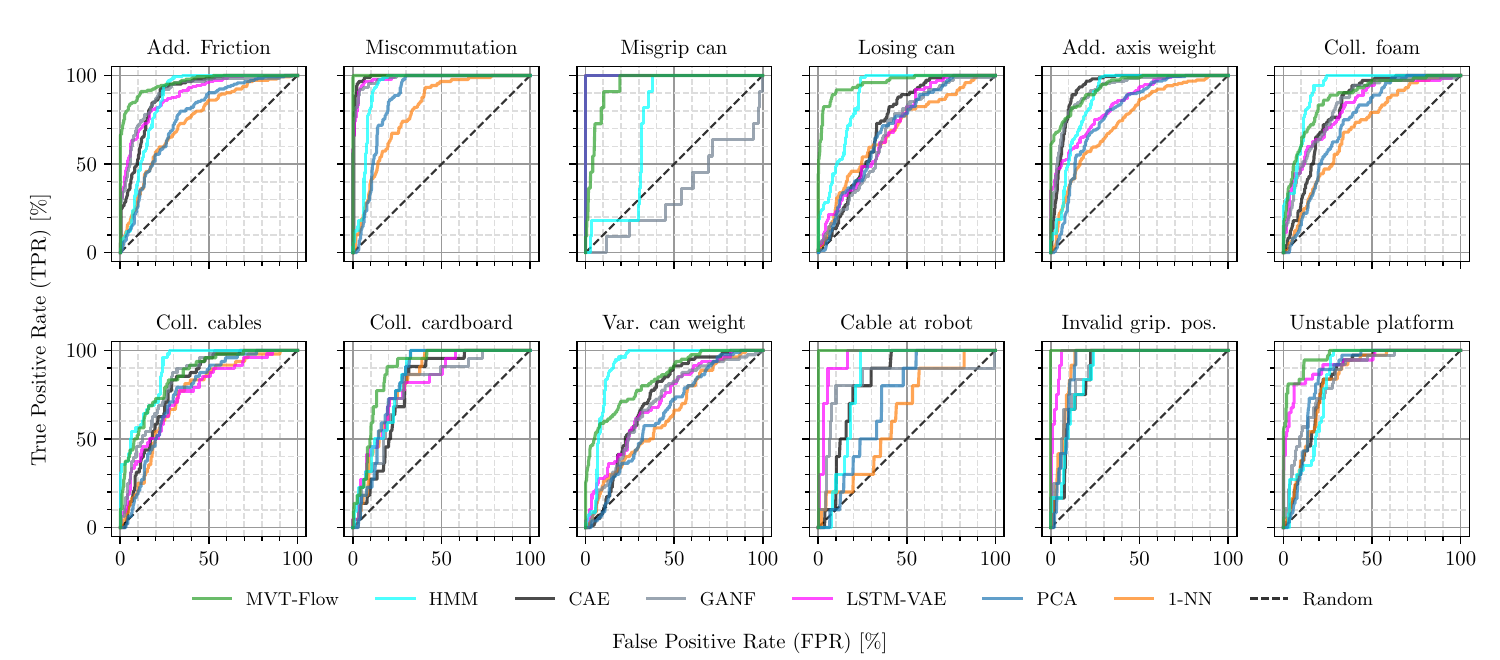}
    \caption{Each diagram shows the ROC-Curve for one anomaly type color-coding the methods.
    }
    \label{fig:analysis_rocs}
    \vspace{-3mm}
    
\end{figure*}

\label{sec:baselines}

\subsubsection{1-NN}
Since the comparison of a sample to its nearest neighbor among normal samples seems to be helpful for humans to identify anomalies (see Fig.~\ref{fig:ano_signals}), we utilize the nearest neighbor distance as anomaly score as proposed by~\cite{amer2012nearest}.
More specifically, we use the $\ell_1$-norm to the nearest neighbor in the training set as a better performance was observed compared to the $\ell_2$-norm.
\subsubsection{PCA}
For the PCA-based approach from~\cite{hornung2014model},
we performed a parameter tuning on the number of components in which 90 principal components explaining 81.1\% of the variance have shown to provide the best performance.
\subsubsection{CAE}
As another baseline, we implemented a convolutional autoencoder (CAE) with the $\ell_2$ reconstruction error as anomaly score, similar to the SWVAE~\cite{swvae} and work from other domains~\cite{chen2018autoencoder, ae_ssim, itae, memae}.
Note that~\cite{swvae} was originally applied for 12 signals (instead of 130 here) having a space complexity of $\mathcal{O}(S^2)$ with $S$ as the number of signals which is why we could not apply it to our dataset for memory reasons.
The CAE includes 1D convolutional layers for the encoder that run over time.
For the decoder, we use transposed convolutions.
A hyperparameter search gives us the following configuration:
Encoder and decoder comprise 3 layers each with a kernel size of 7, strides of 2 and 220 hidden channels.
The latent space has 200 dimensions.
We optimize this network for 60 epochs with Adam~\cite{adam} and an initial learning rate of $5.2 \cdot 10^{-4}$ which is multiplied by $0.1$ after 55 epochs.
{
\subsubsection{GANF}
We used the official repository of Dai and Chen to perform experiments with GANF~\cite{GANF}.
Hyperparameters were taken from their experiments with SWaT~\cite{swat} since the task and dataset is similar to our scenario.
\subsubsection{HMM}
For the HMM method by Azzalini et al.~\cite{hmm_ad} we decided for the online approach as it outperformed the offline approach with less time and space complexity.
As described by the authors, we determined the number of states by the BIC score, running $k$ from 1 to 12, restricting covariance types for the multivariate Gaussian distributed emission probabilities to spherical and diagonal for runtime reasons.
This parameter search yielded the optimal BIC score at k=7 and a covariance with only diagonal entries being nonzero.
\subsubsection{LSTM-VAE}
We used our own reimplementation for the LSTM-VAE from Park et al.~\cite{park2018multimodal}.
Since we observed over-optimization for constant signal sections by having tiny standard deviations in the output, which resulted in poor reconstruction of the remaining signals, we constrained a lower bound of $0.05$ for the standard deviation outputs by adding it to the softplus activation.
It has shown that the LSTM-Memory has problems to process the 130 signals for high sampling rates, which is why we evaluated with 10 Hz here.
A hyperparameter optimization gave us a configuration with 90 epochs, 128 hidden neurons for the LSTMs, a latent space dimension of 32 and a learning rate of $10^{-3}$.}

\label{results}

\begin{figure}
    \centering
    \includegraphics[width=1\linewidth]{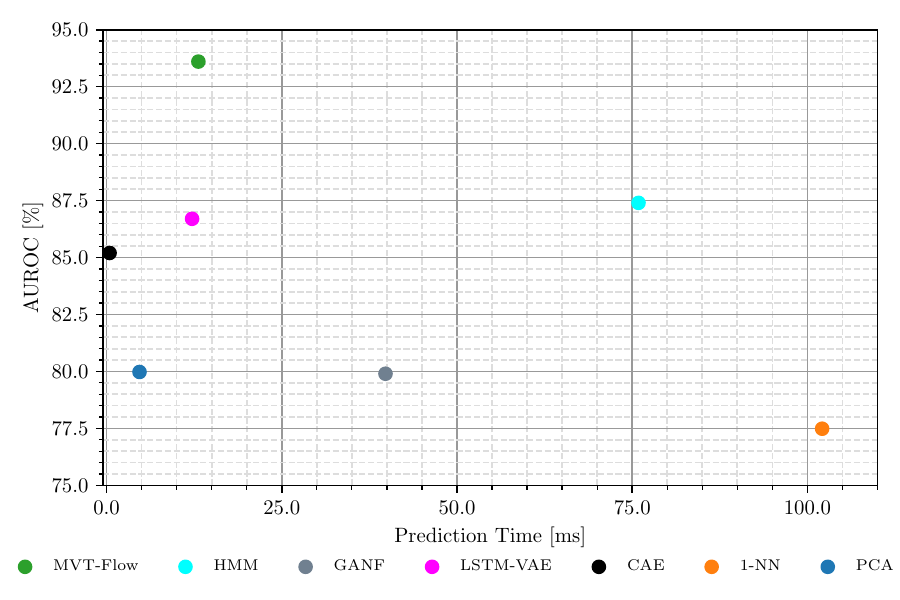}
    \caption{Inference time vs. AD performance for all baselines. {Times for MVT-Flow, GANF, LSTM-VAE and CAE are given using \textit{pytorch} on a \textit{NVIDIA GeForce RTX 4090} as GPU.
     Experiments with HMM, PCA and 1-NN were made with the python packages \textit{hmmlearn}, \textit{pytorch-CPU} and \textit{sklearn}, respectively, and measured with a \textit{Intel i9-13900K} CPU.}}
    \label{fig:durations}
    
\end{figure}
\subsection{Results}
\begin{figure}
    \centering
    \includegraphics[width=1\linewidth]{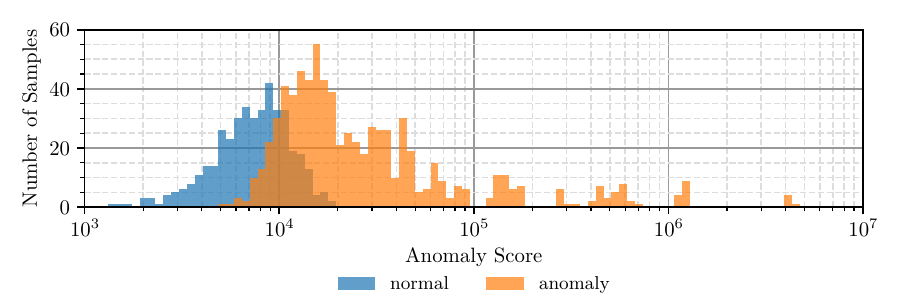}
    \caption{Histogram of anomaly scores for normal data and anomalous data for MVT-Flow. In practice, a threshold inside the overlapping interval around $10^4$ would be chosen.}
    \label{fig:score_distribution}
    \vspace{-3mm}
\end{figure}
\newcommand\ablStudWidth{0.30}

\begin{figure*}[b]
  \centering
  
  \subfloat[][]{\includegraphics[width=\ablStudWidth\textwidth]{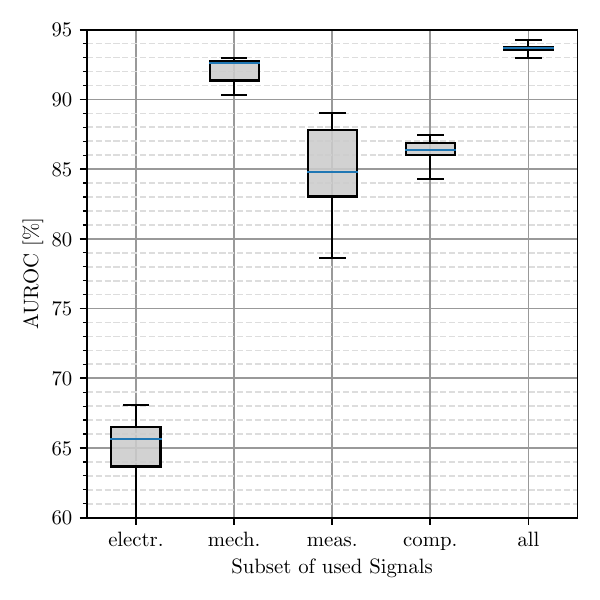} \label{fig:ablation_signals}}%
  \subfloat[][]{\includegraphics[width=\ablStudWidth\textwidth]{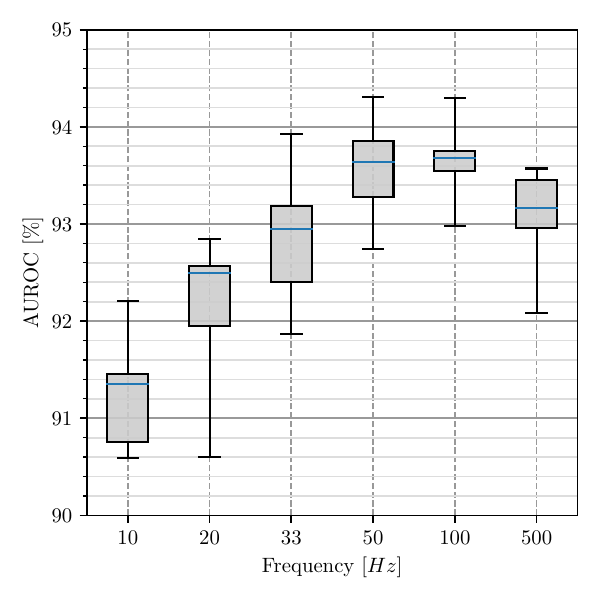} \label{fig:ablation_frequency}}%
  \subfloat[][]{\includegraphics[width=\ablStudWidth\textwidth]{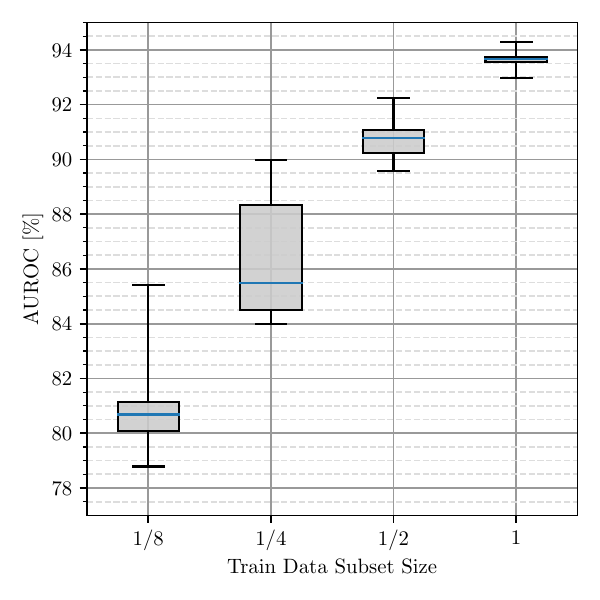} \label{fig:ablation_subsets}}%
  
  \caption{Ablation Studies regarding signal types (a), sampling rate (b) and training set size (c) when applying MVT-Flow on \datasetname{}.}
  \label{fig:ablation}
    \vspace{-6mm}
\end{figure*}

\begin{figure*}[b]
    \centering
    \subfloat[Collision with foam]{\includegraphics[width=0.49\textwidth]{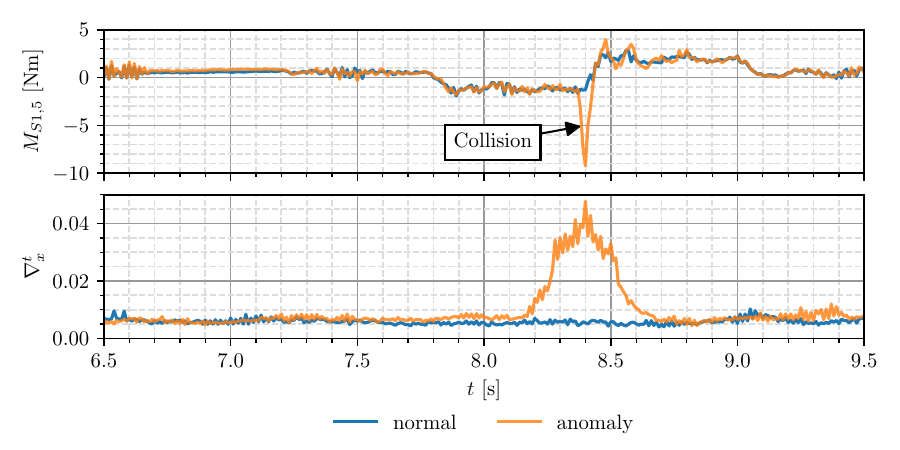}
    \label{fig:temp_analysis_collision}}
    \subfloat[Losing the can]{ \includegraphics[width=0.49\textwidth]{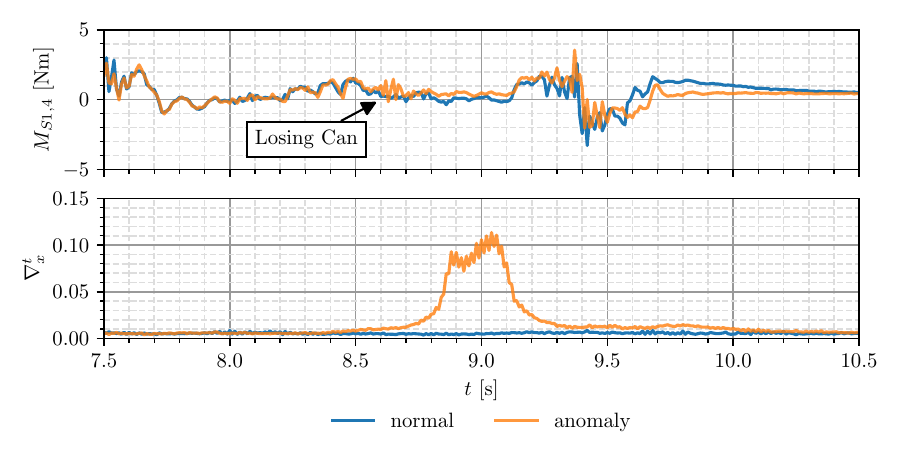}
    \label{fig:temp_analysis_losing}}
    \caption{Temporal analysis: The occurrence of unusual events is temporally determinable by measuring the input gradient. The upper plots show original anomalous signals (orange) and its nearest neighbor among normal samples (blue). The lower plots are the individual input gradients which are highly increased when an unusual event occurs.}
    \label{fig:temp_analysis}
\end{figure*}

In this section, we evaluate current state-of-the-art methods along with MVT-Flow on our new dataset.
The baselines include all the described methods from Sec.~\ref{sec:baselines}.
Table~\ref{table:results} summarizes the results including the AUROC for every anomaly category and the mean AUROC over all of them.
We report the mean and standard deviation over 9 runs {for the non-deterministic trainings of MVT-Flow, LSTM-VAE, GANF, HMM and CAE with different random initializations.}
Figure \ref{fig:analysis_rocs} shows the underlying ROC curves for the median performance over all runs.

{MVT-Flow outperforms all other baselines for 8 out of 12 categories and {on average} by a large margin of 6.2\%.
It is notable that the relative differences in performance between the methods vary strongly depending on the anomaly category.
For example, in contrast to MVT-Flow or HMM, the comparatively simple approaches 1-NN and PCA can perfectly detect a misgrip of the can, but underperform on average.
Hidden Markov Models, on the other hand, detect the misgrip with only 71\% AUROC, but are most sensitive to collisions with about 86 to 92\%.
\newcommand\ms{\text{ }\mathrm{ms}}
Figure~\ref{fig:durations} relates the performance to the inference time of the given methods for samples from our dataset at $100$ Hz.
Although PCA ($4.7\ms$) and CAE ($0.43\ms$) and LSTM-VAE ($12.2\ms$) perform faster than MVT-Flow ($13.1\ms$), all of these methods are fast enough for most real-time requirements in practice.
The nearest neighbor method scales with the training dataset size and might not meet these requirements beyond a certain size (here $102.1 \ms$).
Similarly, the computation time of the HMM approach (here $75.9\ms$) is very sensitive to the number of signals, which affects the computation time of the Hellinger distance.}

The distribution of anomaly scores for normal data and anomalies are visualized in Figure~\ref{fig:score_distribution} on a log scale for anomaly scores.
While the normal samples show almost a normal distribution of rather small scores, we observe a shifted long-tailed distribution towards large scores for anomalies.
Despite there is room for some improvement for many samples in the mid range where the histograms overlap, however, there is a notable ratio of data points which are clearly separated from the opposite class.

\subsection{Ablation Studies}
\label{ablations}
To get a deeper understanding of our dataset and method, we experiment with variations regarding all dimensions including signals, time and sample size.
For all experiments, 9 runs were used with the same random initializations as before.

Figure~\ref{fig:ablation_signals} shows the mean AUROC when using different subsets of signals.
We used the categorization from Table~\ref{tab:signals} and refer to the joint group of targets and estimations as \textit{computed} signals.
Mechanical signals provide clearly more importance for AD compared to electrical signals improving the performance by 25\%.
Both measured and computed signals lead to a moderate performance around 86\% although this varies more for measured signals.
Overall, we observe that each group provides useful information, as the AUROC is above the \textit{random} baseline of 50\% in each case, and that using all signals gives the best performance.

We also investigated the influence of the sample frequency within the range of $10$ to $500$ Hz as shown in Figure~\ref{fig:ablation_frequency}.
{Note that we adapted the dilation parameter for the first convolution of every block to 10 for the $500$ Hz experiment as otherwise a small temporal context would be considered.}
The detection improves with raising frequency {from $10\text{ Hz}$ to $50\text{ Hz}$ as anomalies may be characterized by high-frequent components in this range}.
This effect is mostly saturated {from $50$ Hz and slightly reverses for sampling frequencies up to $500$ Hz.}
Since upcoming methods could still benefit from higher frequencies, we provide signals  with the original sampling at $500$ Hz.

Figure~\ref{fig:ablation_subsets} outlines the relation between training set size and AD performance.
We used random subsets for individual runs {and kept the number of training iterations constant over different subset sizes.}
It turns out that high detection quality is merely achieved with many examples, indicating that detecting anomalies in our dataset is challenging.
Future work may optimize the detection with less data.
\subsection{Temporal Analysis}
We use the input gradient as an indicator for the temporal occurrence of anomalies as described in Sec.~\ref{temp_analysis}.
We evaluate the temporal analysis qualitatively by showing the gradient characteristics for selected examples.
Note that a quantitative evaluation is hardly feasible for many cases since the exact timeframe is indeterminable for gradually increasing or decreasing impacts.

Figure~\ref{fig:temp_analysis} visualizes the torque signal of anomalous sequences and the corresponding input gradient $\nabla_x^t$ from Eq.~\ref{eq:gradient}.
A collision with a foam as shown in  Fig.~\ref{fig:temp_analysis_collision}, which occurs in the original sample {at $8.4$ s} (upper plot) is clearly visible in the gradient (lower plot) at this time.
Note that the receptive field causes some dilatation of this peak around the event.
Interestingly, this peak has a delay for the event \textit{losing the can}, shown in Fig.~\ref{fig:temp_analysis_losing}, since the missing can weight affects the forces in the following movement.
For the anomaly-free sample, the gradient is relatively constant at a low level between $0.05$ and $0.1$ with some noise.
In practice, a simple peak detection after smoothing the signal would provide the unusual parts of the signal.

\section{Conclusion}
We introduce a publicly available AD dataset for robot applications which offers a novel level of variety in anomaly types generalizing to many real scenarios without external sensors.
In addition to the 130 signals of machine data, we provide highly detailed metadata for the pick-and-place cycles.
As a strong baseline, we present MVT-Flow which could transfer the recent success of density estimation with Normalizing Flows to this data domain via a tailored architecture design.
MVT-Flow outperforms previous work regarding sample-wise detection and provides a temporal analysis to inspect unusual events further.

This benchmark makes evaluation of AD methods for robot applications more transparent and comparable such that upcoming methods may demonstrate its effectiveness on this dataset.
In addition to the standard evaluation, we encourage the research community to work on efficiently dealing with the number of signals or samples during training.

{\small
\bibliographystyle{plain}
\bibliography{egbib}

\begin{thebibliography}{10}

\bibitem{amer2012nearest}
Mennatallah Amer and Markus Goldstein.
\newblock Nearest-neighbor and clustering based anomaly detection algorithms for rapidminer.
\newblock In {\em Proc. of the 3rd RapidMiner Community Meeting and Conference (RCOMM 2012)}, pages 1--12, 2012.

\bibitem{cinn}
Lynton Ardizzone, Carsten L{\"u}th, Jakob Kruse, Carsten Rother, and Ullrich K{\"o}the.
\newblock Guided image generation with conditional invertible neural networks.
\newblock {\em arXiv preprint arXiv:1907.02392}, 2019.

\bibitem{azzalini2021minimally}
Davide Azzalini, Luca Bonali, and Francesco Amigoni.
\newblock A minimally supervised approach based on variational autoencoders for anomaly detection in autonomous robots.
\newblock {\em IEEE Robotics and Automation Letters}, 6(2):2985--2992, 2021.

\bibitem{hmm_ad}
Davide Azzalini, Alberto Castellini, Matteo Luperto, Alessandro Farinelli, Francesco Amigoni, et~al.
\newblock {HMMs} for anomaly detection in autonomous robots.
\newblock In {\em AAMAS CONFERENCE PROCEEDINGS}, pages 105--113. ACM, 2020.

\bibitem{bahrin2016industry}
Mohd Aiman~Kamarul Bahrin, Mohd~Fauzi Othman, Nor Hayati~Nor Azli, and Muhamad~Farihin Talib.
\newblock Industry 4.0: A review on industrial automation and robotic.
\newblock {\em Jurnal teknologi}, 78(6-13), 2016.

\bibitem{hmm}
Leonard~E Baum and Ted Petrie.
\newblock Statistical inference for probabilistic functions of finite state markov chains.
\newblock {\em The annals of mathematical statistics}, 37(6):1554--1563, 1966.

\bibitem{st_bergmann2}
Paul Bergmann, Kilian Batzner, Michael Fauser, David Sattlegger, and Carsten Steger.
\newblock Beyond dents and scratches: Logical constraints in unsupervised anomaly detection and localization.
\newblock {\em Int. J. Comput. Vis.}, 130(4):947--969, 2022.

\bibitem{mvtec}
Paul Bergmann, Michael Fauser, David Sattlegger, and Carsten Steger.
\newblock {MVTec} ad--a comprehensive real-world dataset for unsupervised anomaly detection.
\newblock In {\em Proceedings of the IEEE Conference on Computer Vision and Pattern Recognition}, pages 9592--9600, 2019.

\bibitem{mvtec3d}
Paul Bergmann, Xin Jin, David Sattlegger, and Carsten Steger.
\newblock The {MVTec 3D-AD} dataset for unsupervised {3D} anomaly detection and localization.
\newblock In Giovanni~Maria Farinella, Petia Radeva, and Kadi Bouatouch, editors, {\em Proceedings of the 17th International Joint Conference on Computer Vision, Imaging and Computer Graphics Theory and Applications, {VISIGRAPP} 2022, Volume 5: VISAPP, Online Streaming, February 6-8, 2022}, pages 202--213. {SCITEPRESS}, 2022.

\bibitem{ae_ssim}
Paul Bergmann, Sindy L{\"o}we, Michael Fauser, David Sattlegger, and C.~Steger.
\newblock Improving unsupervised defect segmentation by applying structural similarity to autoencoders.
\newblock In {\em VISIGRAPP}, 2019.

\bibitem{swvae}
Tingting Chen, Xueping Liu, Bizhong Xia, Wei Wang, and Yongzhi Lai.
\newblock Unsupervised anomaly detection of industrial robots using sliding-window convolutional variational autoencoder.
\newblock {\em IEEE Access}, 8:47072--47081, 2020.

\bibitem{chen2018autoencoder}
Zhaomin Chen, Chai~Kiat Yeo, Bu~Sung Lee, and Chiew~Tong Lau.
\newblock Autoencoder-based network anomaly detection.
\newblock In {\em 2018 Wireless telecommunications symposium (WTS)}, pages 1--5. IEEE, 2018.

\bibitem{GANF}
Enyan Dai and Jie Chen.
\newblock Graph-augmented normalizing flows for anomaly detection of multiple time series.
\newblock In {\em The Tenth International Conference on Learning Representations, {ICLR} 2022, Virtual Event, April 25-29, 2022}. OpenReview.net, 2022.

\bibitem{nf_trajectory}
Madson L.~D. Dias, C{\'{e}}sar Lincoln~C. Mattos, Ticiana L.~Coelho da~Silva, Jos{\'{e}} Ant{\^{o}}nio~Fernandes de~Mac{\^{e}}do, and Wellington C.~P. Silva.
\newblock Anomaly detection in trajectory data with normalizing flows.
\newblock In {\em 2020 International Joint Conference on Neural Networks, {IJCNN} 2020, Glasgow, United Kingdom, July 19-24, 2020}, pages 1--8. {IEEE}, 2020.

\bibitem{realnvp}
Laurent Dinh, Jascha Sohl-Dickstein, and Samy Bengio.
\newblock Density estimation using {Real-NVP}.
\newblock {\em ICLR 2017}, 2016.

\bibitem{itae}
Ye~Fei, Chaoqin Huang, Cao Jinkun, Maosen Li, Ya~Zhang, and Cewu Lu.
\newblock Attribute restoration framework for anomaly detection.
\newblock {\em IEEE Transactions on Multimedia}, 2020.

\bibitem{germain}
Mathieu Germain, Karol Gregor, Iain Murray, and Hugo Larochelle.
\newblock Made: Masked autoencoder for distribution estimation.
\newblock In {\em International Conference on Machine Learning}, pages 881--889, 2015.

\bibitem{swat}
Jonathan Goh, Sridhar Adepu, Khurum~Nazir Junejo, and Aditya Mathur.
\newblock A dataset to support research in the design of secure water treatment systems.
\newblock In {\em Critical Information Infrastructures Security: 11th International Conference, CRITIS 2016, Paris, France, October 10--12, 2016, Revised Selected Papers 11}, pages 88--99. Springer, 2017.

\bibitem{memae}
Dong Gong, Lingqiao Liu, Vuong Le, Budhaditya Saha, Moussa~Reda Mansour, Svetha Venkatesh, and Anton van~den Hengel.
\newblock Memorizing normality to detect anomaly: Memory-augmented deep autoencoder for unsupervised anomaly detection.
\newblock In {\em Proceedings of the IEEE International Conference on Computer Vision}, pages 1705--1714, 2019.

\bibitem{gan}
Ian Goodfellow, Jean Pouget-Abadie, Mehdi Mirza, Bing Xu, David Warde-Farley, Sherjil Ozair, Aaron Courville, and Yoshua Bengio.
\newblock Generative adversarial nets.
\newblock In {\em Advances in neural information processing systems}, pages 2672--2680, 2014.

\bibitem{ffjord}
Will Grathwohl, Ricky~TQ Chen, Jesse Bettencourt, Ilya Sutskever, and David Duvenaud.
\newblock {FFJORD}: Free-form continuous dynamics for scalable reversible generative models.
\newblock In {\em International Conference on Learning Representations 2019}.

\bibitem{cflow}
Denis Gudovskiy, Shun Ishizaka, and Kazuki Kozuka.
\newblock {CFLOW-AD}: Real-time unsupervised anomaly detection with localization via conditional normalizing flows.
\newblock In {\em Proceedings of the IEEE/CVF Winter Conference on Applications of Computer Vision}, pages 98--107, 2022.

\bibitem{hellinger1909neue}
Ernst Hellinger.
\newblock Neue begr{\"u}ndung der theorie quadratischer formen von unendlichvielen ver{\"a}nderlichen.
\newblock {\em Journal f{\"u}r die reine und angewandte Mathematik}, 1909(136):210--271, 1909.

\bibitem{hornung2014model}
Rachel Hornung, Holger Urbanek, Julian Klodmann, Christian Osendorfer, and Patrick Van Der~Smagt.
\newblock Model-free robot anomaly detection.
\newblock In {\em 2014 IEEE/RSJ International Conference on Intelligent Robots and Systems}, pages 3676--3683. IEEE, 2014.

\bibitem{kang2022traffic}
Zhuangwei Kang, Ayan Mukhopadhyay, Aniruddha Gokhale, Shijie Wen, and Abhishek Dubey.
\newblock Traffic anomaly detection via conditional normalizing flow.
\newblock In {\em 2022 IEEE 25th International Conference on Intelligent Transportation Systems (ITSC)}, pages 2563--2570. IEEE, 2022.

\bibitem{khalastchi2018fault}
Eliahu Khalastchi and Meir Kalech.
\newblock On fault detection and diagnosis in robotic systems.
\newblock {\em ACM Computing Surveys (CSUR)}, 51(1):1--24, 2018.

\bibitem{khalastchi2015online}
Eliahu Khalastchi, Meir Kalech, Gal~A Kaminka, and Raz Lin.
\newblock Online data-driven anomaly detection in autonomous robots.
\newblock {\em Knowledge and Information Systems}, 43(3):657--688, 2015.

\bibitem{adam}
Diederik~P Kingma and Jimmy Ba.
\newblock Adam: A method for stochastic optimization.
\newblock In {\em International Conference on Learning Representations (ICLR)}, 2015.

\bibitem{vae}
Diederik~P. Kingma and Max Welling.
\newblock Auto-encoding variational bayes.
\newblock {\em CoRR}, abs/1312.6114, 2013.

\bibitem{kingma2018glow}
Durk~P Kingma and Prafulla Dhariwal.
\newblock Glow: Generative flow with invertible 1x1 convolutions.
\newblock {\em Advances in neural information processing systems}, 31, 2018.

\bibitem{kingma}
Durk~P Kingma, Tim Salimans, Rafal Jozefowicz, Xi~Chen, Ilya Sutskever, and Max Welling.
\newblock Improved variational inference with inverse autoregressive flow.
\newblock In {\em Advances in neural information processing systems}, pages 4743--4751, 2016.

\bibitem{kobyzev2020normalizing}
Ivan Kobyzev, Simon~JD Prince, and Marcus~A Brubaker.
\newblock Normalizing flows: An introduction and review of current methods.
\newblock {\em IEEE transactions on pattern analysis and machine intelligence}, 43(11):3964--3979, 2020.

\bibitem{kragic2018interactive}
Danica Kragic, Joakim Gustafson, Hakan Karaoguz, Patric Jensfelt, and Robert Krug.
\newblock Interactive, collaborative robots: Challenges and opportunities.
\newblock In {\em IJCAI}, pages 18--25, 2018.

\bibitem{aursad}
B{\l}a{\.z}ej Leporowski, Daniella Tola, Casper Hansen, and Alexandros Iosifidis.
\newblock Detecting faults during automatic screwdriving: A dataset and use case of anomaly detection for automatic screwdriving.
\newblock In {\em Towards Sustainable Customization: Bridging Smart Products and Manufacturing Systems}, pages 224--232. Springer, 2021.

\bibitem{ad_survey1}
Guansong Pang, Chunhua Shen, Longbing Cao, and Anton Van~Den Hengel.
\newblock Deep learning for anomaly detection: A review.
\newblock {\em ACM Computing Surveys (CSUR)}, 54(2):1--38, 2021.

\bibitem{papamakarios2021normalizing}
George Papamakarios, Eric Nalisnick, Danilo~Jimenez Rezende, Shakir Mohamed, and Balaji Lakshminarayanan.
\newblock Normalizing flows for probabilistic modeling and inference.
\newblock {\em The Journal of Machine Learning Research}, 22(1):2617--2680, 2021.

\bibitem{maf}
George Papamakarios, Theo Pavlakou, and Iain Murray.
\newblock Masked autoregressive flow for density estimation.
\newblock {\em Advances in neural information processing systems}, 30, 2017.

\bibitem{park2016multimodal}
Daehyung Park, Zackory Erickson, Tapomayukh Bhattacharjee, and Charles~C Kemp.
\newblock Multimodal execution monitoring for anomaly detection during robot manipulation.
\newblock In {\em 2016 IEEE International Conference on Robotics and Automation (ICRA)}, pages 407--414. IEEE, 2016.

\bibitem{park2018multimodal}
Daehyung Park, Yuuna Hoshi, and Charles~C Kemp.
\newblock A multimodal anomaly detector for robot-assisted feeding using an lstm-based variational autoencoder.
\newblock {\em IEEE Robotics and Automation Letters}, 3(3):1544--1551, 2018.

\bibitem{park2019multimodal}
Daehyung Park, Hokeun Kim, and Charles~C Kemp.
\newblock Multimodal anomaly detection for assistive robots.
\newblock {\em Autonomous Robots}, 43:611--629, 2019.

\bibitem{pca}
Karl Pearson.
\newblock Liii. on lines and planes of closest fit to systems of points in space.
\newblock {\em The London, Edinburgh, and Dublin philosophical magazine and journal of science}, 2(11):559--572, 1901.

\bibitem{nf}
Danilo Rezende and Shakir Mohamed.
\newblock Variational inference with normalizing flows.
\newblock In {\em International Conference on Machine Learning}, pages 1530--1538. PMLR, 2015.

\bibitem{romeres2019anomaly}
Diego Romeres, Devesh~K Jha, William Yerazunis, Daniel Nikovski, and Hoang~Anh Dau.
\newblock Anomaly detection for insertion tasks in robotic assembly using gaussian process models.
\newblock In {\em 2019 18th European Control Conference (ECC)}, pages 1017--1022. IEEE, 2019.

\bibitem{differnet}
Marco Rudolph, Bastian Wandt, and Bodo Rosenhahn.
\newblock Same same but differnet: Semi-supervised defect detection with normalizing flows.
\newblock In {\em Proceedings of the IEEE/CVF Winter Conference on Applications of Computer Vision}, pages 1907--1916, 2021.

\bibitem{csflow}
Marco Rudolph, Tom Wehrbein, Bodo Rosenhahn, and Bastian Wandt.
\newblock Fully convolutional cross-scale-flows for image-based defect detection.
\newblock In {\em Proceedings of the IEEE/CVF Winter Conference on Applications of Computer Vision}, pages 1088--1097, 2022.

\bibitem{ast}
Marco Rudolph, Tom Wehrbein, Bodo Rosenhahn, and Bastian Wandt.
\newblock Asymmetric student-teacher networks for industrial anomaly detection.
\newblock In {\em Proceedings of the IEEE/CVF Winter Conference on Applications of Computer Vision}, pages 2592--2602, 2023.

\bibitem{ad_review1}
Lukas Ruff, Jacob~R Kauffmann, Robert~A Vandermeulen, Gr{\'e}goire Montavon, Wojciech Samek, Marius Kloft, Thomas~G Dietterich, and Klaus-Robert M{\"u}ller.
\newblock A unifying review of deep and shallow anomaly detection.
\newblock {\em Proceedings of the IEEE}, 109(5):756--795, 2021.

\bibitem{nf_deep}
Artem Ryzhikov, Maxim Borisyak, Andrey Ustyuzhanin, and Denis Derkach.
\newblock Normalizing flows for deep anomaly detection.
\newblock {\em arXiv preprint arXiv:1912.09323}, 2019.

\bibitem{nf_time_series}
Maximilian Schmidt and Marko Simic.
\newblock Normalizing flows for novelty detection in industrial time series data.
\newblock {\em arXiv preprint arXiv:1906.06904}, 2019.

\bibitem{rnn_vae}
Maximilian S{\"o}lch, Justin Bayer, Marvin Ludersdorfer, and Patrick van~der Smagt.
\newblock Variational inference for on-line anomaly detection in high-dimensional time series.
\newblock {\em arXiv preprint arXiv:1602.07109}, 2016.

\bibitem{ad_survey2}
Srikanth Thudumu, Philip Branch, Jiong Jin, and Jugdutt~Jack Singh.
\newblock A comprehensive survey of anomaly detection techniques for high dimensional big data.
\newblock {\em Journal of Big Data}, 7(1):1--30, 2020.

\bibitem{vasic2013safety}
Milos Vasic and Aude Billard.
\newblock Safety issues in human-robot interactions.
\newblock In {\em 2013 ieee international conference on robotics and automation}, pages 197--204. IEEE, 2013.

\bibitem{tomINN}
Tom Wehrbein, Marco Rudolph, Bodo Rosenhahn, and Bastian Wandt.
\newblock Probabilistic monocular {3D} human pose estimation with normalizing flows.
\newblock {\em Proceedings of the IEEE International Conference on Computer Vision}, 2021.

\bibitem{yan2022cainnflow}
Ruiqing Yan, Fan Zhang, Mengyuan Huang, Wu~Liu, Dongyu Hu, Jinfeng Li, Qiang Liu, Jingrong Jiang, Qianjin Guo, and Linghan Zheng.
\newblock Cainnflow.
\newblock {\em arXiv preprint arXiv:2206.01992}, 2022.

\bibitem{zhang2021robot}
Tie Zhang, Peizhong Ge, Yanbiao Zou, and Yingwu He.
\newblock Robot collision detection without external sensors based on time-series analysis.
\newblock {\em Journal of Dynamic Systems, Measurement, and Control}, 143(4), 2021.

\end{thebibliography}
}

\newcommand\vspacebio{\vspace{-1cm}}

\vspacebio{}

\begin{IEEEbiography}[{\includegraphics[trim=210 200 210 80,width=1in,height=1.25in,clip,keepaspectratio]
		{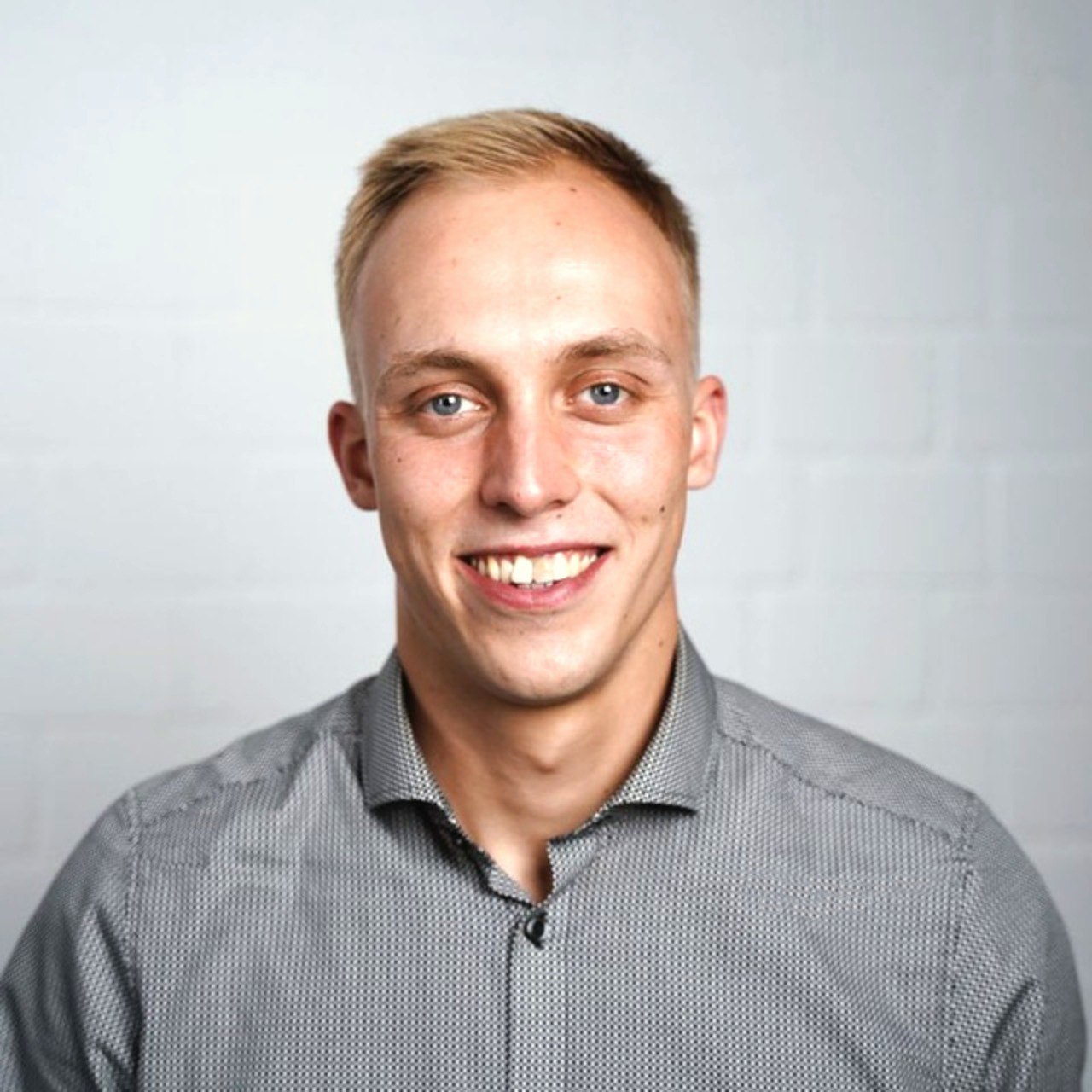}}]{Jan Thieß Brockmann}
	 received his B.Sc. and M.Sc. in electrical engineering and information technology from the Leibniz University of Hanover in 2020 and 2022, respectively. His master thesis is the base of the here published work. Since 2022 he works for voraus robotik GmbH in Hanover as a software and cloud engineer.
\end{IEEEbiography}

\vspacebio{}

\begin{IEEEbiography}[{\includegraphics[trim=0 20 0 0, width=1in,height=1.25in,clip,keepaspectratio]
		{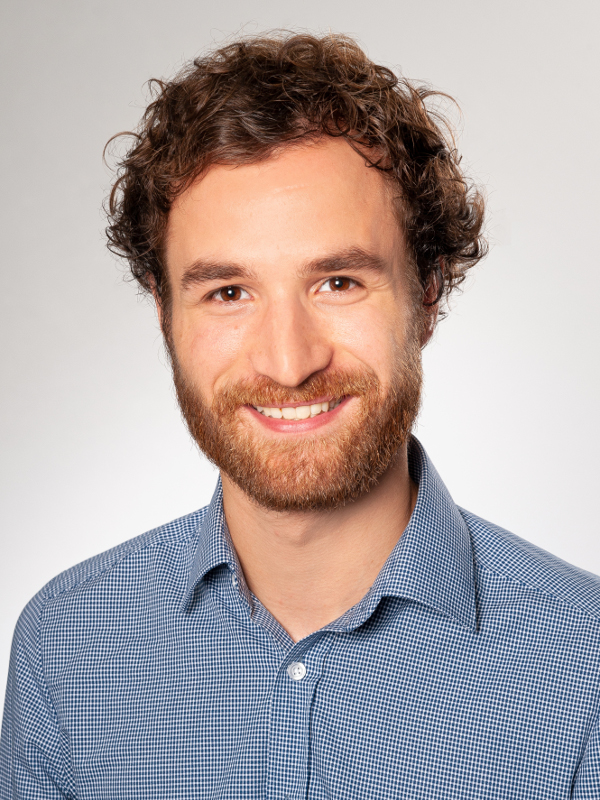}}]{Marco Rudolph}
	 received his B.Sc. and M.Sc. in computer science from the Leibniz University of Hannover in 2016 and 2018, respectively. Since 10/2018 he works as a research assistant
towards his PhD in the group of Prof. Rosenhahn. His
research interests are in the fields of anomaly detection, interpretable machine learning and human pose estimation, mostly specialized on image data.
\end{IEEEbiography}

\vspacebio{}

\begin{IEEEbiography}[{\includegraphics[trim=0 30 0 0, width=1in,height=1.25in,clip,keepaspectratio]
		{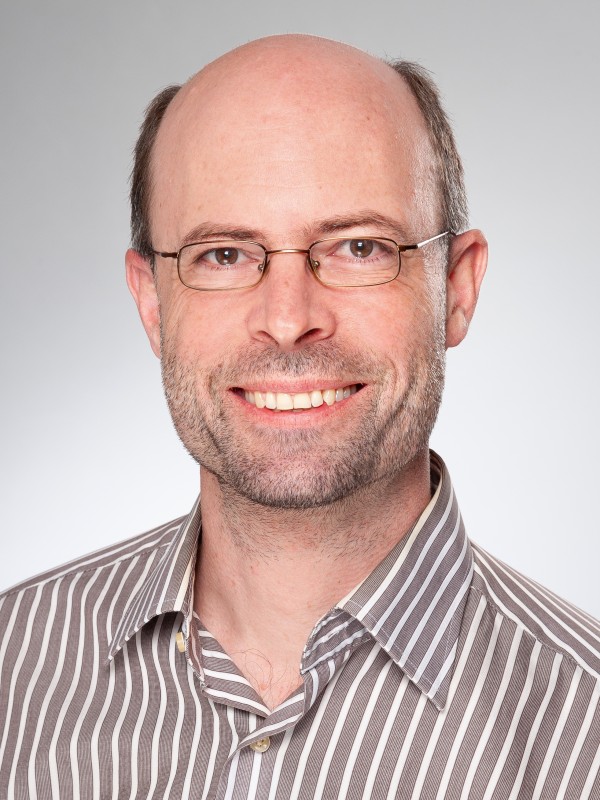}}]{Bodo Rosenhahn}
	 received the Dipl.-Inf. and Dr.-Ing.
degrees from the University of Kiel in 1999 and 2003,
respectively. He studied computer science (minor
subject Medicine) with the University of Kiel. From
10/2003 till 10/2005, he worked as PostDoc with the
University of Auckland (New Zealand). In 11/2005-08/2008 he worked as senior Researcher with the Max-Planck Institute for Computer
Science. Since 09/2008 he is full professor with the
Leibniz University of Hannover, heading a group on
automated image interpretation. He has co-authored more than 400 papers, holds
12 patents and organized several workshops and conferences in the last years.
His works received several awards, including a DAGM-Prize 2002, the Dr.-Ing.
Siegfried Werth Prize 2003, the DAGM-Main Prize 2005 and
2007, the Olympus-Prize 2007, and the Günter Enderle Award (Eurographics)
2017.
\end{IEEEbiography}

\vspacebio{}

\begin{IEEEbiography}[{\includegraphics[width=1in,height=1.25in,clip,keepaspectratio]
		{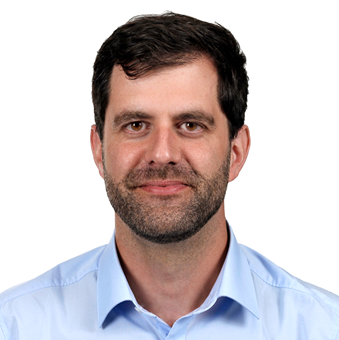}}]{Bastian Wandt}
	 received his M.Sc. and Dr.-Ing. degree from the Leibniz University of Hannover in 2014 and 2020. From 01/2021 to 09/2022 he worked as a PostDoc at the University of British Columbia in Vancouver, Canada. His PostDoc research was partially funded by the German Research Foundation (DFG) under the Walter Benjamin scholarship. Since September 2022 he is an assistant professor at the Linköping University, Sweden.
\end{IEEEbiography}

\end{document}